\documentclass[nohyperref]{article}
\usepackage[utf8]{inputenc} 
\usepackage[T1]{fontenc}    
\usepackage{microtype}
\usepackage{times}
\usepackage{graphicx}
\usepackage{amsfonts,amsmath,amssymb,amsthm,mathtools,systeme}
\usepackage{bm}
\usepackage{acronym}
\usepackage{enumitem}
\usepackage[pagebackref,breaklinks,colorlinks]{hyperref}
\usepackage{xspace}
\usepackage{setspace}
\usepackage{subfigure}
\usepackage[skip=0pt,font=small]{caption}
\usepackage[dvipsnames]{xcolor}
\usepackage[capitalise]{cleveref}
\usepackage{booktabs,tabularx,colortbl,multirow,array,makecell}

\makeatletter
\DeclareRobustCommand\onedot{\futurelet\@let@token\@onedot}
\def\@onedot{\ifx\@let@token.\else.\null\fi\xspace}
\def\eg{\emph{e.g}\onedot} 

\def\ie{\emph{i.e}\onedot}

\makeatother

\DeclareMathOperator{\ELBO}{\mathrm{ELBO}}
\DeclareMathOperator{\KL}{\mathbb{D}_{\rm KL} }
\DeclareMathOperator{\HH}{\mathcal{H}}
\DeclareMathOperator{\II}{\mathcal{I}}

\newcommand{\x}{\mathbf{x}}
\newcommand{\y}{\mathbf{y}}
\newcommand{\z}{\mathbf{z}}
\newcommand{\I}{\mathbf{I}}
\newcommand{\E}{\mathbb{E}}
\newcommand{\R}{\mathbb{R}}
\newcommand{\N}{\mathcal{N}}

\makeatletter
\renewcommand{\paragraph}{%
  \@startsection{paragraph}{4}%
  {\z@}{0ex \@plus 0ex \@minus 0ex}{-1em}%
  {\hskip\parindent\normalfont\normalsize\bfseries}%
}
\makeatother

\graphicspath{{figures/}}

\crefname{algorithm}{Alg.}{Algs.}
\Crefname{algocf}{Algorithm}{Algorithms}
\crefname{section}{Sec.}{Secs.}
\Crefname{section}{Section}{Sections}
\crefname{table}{Tab.}{Tabs.}
\Crefname{table}{Table}{Tables}
\crefname{figure}{Fig.}{Fig.}
\Crefname{figure}{Figure}{Figure}
\crefname{appendix}{Appx.}{Appx.}
\Crefname{appendix}{Appendix}{Appendix}

\definecolor{gblue}{HTML}{4285F4}
\definecolor{gred}{HTML}{DB4437}
\definecolor{ggreen}{HTML}{0F9D58}
\definecolor{gray}{gray}{0.9}
\definecolor{tgray}{gray}{0.5}

\usepackage{tikz}
\usetikzlibrary{positioning, arrows.meta, quotes}
\usetikzlibrary{shapes,decorations}
\usetikzlibrary{bayesnet}
\tikzset{>=latex}
\tikzstyle{plate caption} = [
    caption, node distance=0, inner sep=0pt,
    below left=5pt and 0pt of #1.south
]
\tikzstyle{nbase} = [
    circle, 
    draw=black, 
    fill=white, 
    inner sep=0pt, 
    minimum size=0.8cm,
    node distance=16mm
]

\acrodef{dlvm}[DLVM]{Deep Latent Variable Model}
\acrodef{vae}[VAE]{Variational Auto-Encoder}
\acrodef{ebm}[EBM]{Energy-Based Model}
\acrodef{svebm}[SVEBM]{Symbol-Vector Coupling Energy-Based Model}
\acrodef{ldebm}[LDEBM]{Latent Diffusion Energy-Based Model}
\acrodef{ib}[IB]{Information Bottleneck}
\acrodef{gc}[GC]{Geometric Clustering}
\acrodef{kde}[KDE]{Kernel Density Estimation}
\acrodef{rppl}[rPPL]{Reverse Perplexity}
\acrodef{wkl}[wKL]{Word-Level KL Divergence}
\acrodef{nll}[NLL]{Negative Log-Likelihood}
\acrodef{ptb}[PTB]{Penn Treebanks}
\acrodef{dd}[DD]{Daily Dialog}
\acrodef{smd}[SMD]{Stanford Multi-Domain Dialog}
\acrodef{mcmc}[MCMC]{Markov Chain Monte Carlo}

\newcolumntype{C}{>{\columncolor{gray}}c}
\newcolumntype{L}{>{\columncolor{gray}}l}
\newcommand{\tabincell}[2]{\begin{tabular}{@{}#1@{}}#2\end{tabular}}

\usepackage[accepted]{icml2022}

\icmltitlerunning{Latent Diffusion Energy-Based Model for Interpretable Text Modeling}

\begin{document}

\twocolumn[
\icmltitle{Latent Diffusion Energy-Based Model for Interpretable Text Modeling}



\icmlsetsymbol{equal}{*}

\begin{icmlauthorlist}
\icmlauthor{Peiyu Yu}{ucla_cs,bigai}
\icmlauthor{Sirui Xie}{ucla_cs}
\icmlauthor{Xiaojian Ma}{ucla_cs,bigai}
\icmlauthor{Baoxiong Jia}{ucla_cs,bigai}
\icmlauthor{Bo Pang}{salesforce}\\
\icmlauthor{Ruiqi Gao}{google}
\icmlauthor{Yixin Zhu}{pku_iai,pku_sai}
\icmlauthor{Song-Chun Zhu}{ucla_cs,bigai,pku_iai,pku_sai,ucla_stats,thu}
\icmlauthor{Ying Nian Wu}{ucla_stats}
\end{icmlauthorlist}

\icmlaffiliation{ucla_cs}{Department of Computer Science, UCLA, USA}
\icmlaffiliation{ucla_stats}{Department of Statistics, UCLA, USA}
\icmlaffiliation{salesforce}{Salesforce Research, USA}
\icmlaffiliation{google}{Google Brain, USA}
\icmlaffiliation{bigai}{Beijing Institute for General Artificial Intelligence, China}
\icmlaffiliation{pku_iai}{Institute for Artificial Intelligence, Peking University, China}
\icmlaffiliation{pku_sai}{School of Artificial Intelligence, Peking University, China}
\icmlaffiliation{thu}{Department of Automation, Tsinghua University, China}

\icmlcorrespondingauthor{Peiyu Yu}{yupeiyu98@g.ucla.edu}

\icmlkeywords{Energy-Based Model}
\vskip 0.3in
]



\printAffiliationsAndNotice{Code repo and data: \href{https://github.com/yuPeiyu98/Latent-Diffusion-EBM}{https://github.com/yuPeiyu98/Latent-Diffusion-EBM}.}

\begin{abstract}
Latent space \acp{ebm}, also known as energy-based priors, have drawn growing interests in generative modeling. Fueled by its flexibility in the formulation and strong modeling power of the latent space, recent works built upon it have made interesting attempts aiming at the interpretability of text modeling. However, latent space \acp{ebm} also inherit some flaws from \acp{ebm} in data space; the degenerate \acs{mcmc} sampling quality in practice can lead to poor generation quality and instability in training, especially on data with complex latent structures. Inspired by the recent efforts that leverage diffusion recovery likelihood learning as a cure for the sampling issue, we introduce a novel symbiosis between the diffusion models and latent space \acp{ebm} in a variational learning framework, coined as the \textit{latent diffusion energy-based model}. We develop a geometric clustering-based regularization jointly with the information bottleneck to further improve the quality of the learned latent space. Experiments on several challenging tasks demonstrate the superior performance of our model on interpretable text modeling over strong counterparts.
\end{abstract}

\section{Introduction}\label{sec:intro}

Text modeling has achieved impressive progress with the fast development of neural generative models \citep{serban2016building,li2017deep,zhao2017learning,gupta2018deep,zhao2018adversarially}. It allows near human-level text generation quality and also leads to a wide range of real-world applications such as dialog system \cite{young2013pomdp} and machine translation \cite{brown1993mathematics}. Although the quality of generation (\eg, fluency and diversity) is the primary concern of most work, interpretability of the generation process has drawn much attention recently. Among the existing frameworks, the \ac{dlvm} is especially suitable for the task, as the learned latent space could capture high-level structures with semantic meanings like topics \citep{wang2019topic} and dialog actions \citep{zhao2018unsupervised}; such latent space could further enable more interpretable text modeling, featuring unsupervised text attributes discovery \citep{wen2017latent}, conditional and controllable text generation \citep{fang2019implicit,shi2020dispersed}, and semi-supervised text classification \citep{pang2021latent}.

\begin{figure}[t!]
    \centering
    \begin{tikzpicture}
        \node [nbase, fill=gray] (x) at (-3.2,0) {$\x$};
        \node [nbase] (z0) at (-0.75,0) {$\z_0$};
        \node [nbase] (zt) at (1,0) {$\z_t$};
        \node [nbase] (zt1) at (3.25,0) {$\z_{t+1}$};
        \node [nbase] (y) at (-0.75,1.5) {$\y$};
    
        \node (forward) at (2.2, 0.5) 
            {$q(\z_{t+1}|\z_t)$};
        \node (backward) at (2.2, -0.7) 
            {$p_\alpha(\z_t|\z_{t+1})$};
        \node (couple) at (0.8, 1.5)
            {$p_\alpha (\y, \z_0|\z_1)$};
        \node (encode) at (-2.0, 0.5)
            {$q_\phi(\z_0|\x)$};
        \node (decode) at (-2.0, -0.7)
            {$p_\beta(\x|\z_0)$};
        
        \path [draw, ->] (x) edge (z0);
        \path [draw, ->, dashed] 
            (z0) edge [bend left] node [right] {} (x);
        \path [draw, <->] (y) edge (z0);
        \path [draw, ->, color=blue!50!green] (z0) edge (zt);
        \path [draw, ->, dashed, color=blue!50!green] 
            (zt) edge [bend left] node [right] {} (z0);
        \path [draw, ->] (zt) edge (zt1);
        \path [draw, ->, dashed] 
            (zt1) edge [bend left] node [right] {} (zt);
    
        \plate [color=blue] {part1} {(zt)(zt1)(forward)(backward)} 
            {$t=1,...,T-1$};
        \plate [color=red] {part2} {(y)(z0)(couple)(part1)} {};
    \end{tikzpicture}
    \caption{\textbf{Graphical illustration of the latent diffusion process.} We construct the forward and reverse diffusion processes in the latent space. The symbolic one-hot vector is coupled with the initial latent vector $\z_0$. The latent and diffused latent variables are highlighted by the red and blue plates, respectively. The cyan arrows indicate that $\z_0$ is connected with only $\z_1$. We learn a sequence of \acsp{ebm} to model the reverse diffusion process $p_\alpha(\z_t|\z_{t+1})$.}
    \label{fig:pgm}
\end{figure}
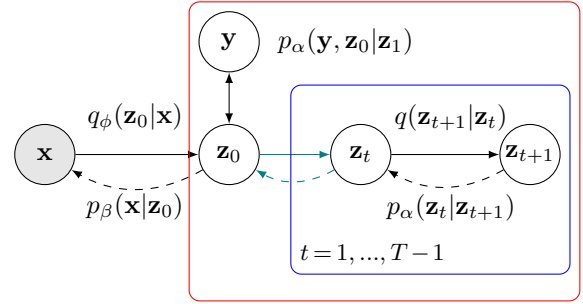

In essence, \ac{dlvm} summarizes the observed sample (\eg, a piece of text) into inferred latent variables. Earlier text-modeling methods with \ac{dlvm} mostly follow the formulation of \ac{vae} \citep{kingma2013auto,rezende2014stochastic,bowman2016generating}, which assumes a continuous latent space. More recently, \citet{zhao2018unsupervised} explore the possibility of using a discrete latent space to capture dialog actions; \citet{shi2020dispersed} propose to use \ac{vae} with the mixture of Gaussians as the prior, demonstrating promising interpretability of dialog utterance generation. To further improve the expressivity of the latent space, \citet{pang2021latent} leverage the flexibility of \textit{energy-based prior} \citep{pang2020learning} and learn a structured latent space for interpretable text generation and classification. Specifically, they propose a symbol-vector coupling prior model. The continuous latent variables are coupled with discrete one-hot symbol variables, allowing better discrete structure induction without sacrificing the generation quality offered by the continuous latent space. However, similar to learning an \ac{ebm} in data space, the learning of energy-based prior requires \ac{mcmc} sampling, whose quality can degenerate in practice \citep{grathwohl2019your,nijkamp2019learning,nijkamp2020anatomy,gao2020learning}, especially on data with complex latent structures; it often leads to instability during training. As we demonstrate empirically in \cref{sec:gen_model}, this phenomenon is particularly concerning when adopting the variational learning scheme to update model parameters.

To remedy this \ac{mcmc} sampling issue, we may take a look at the endeavor of \ac{ebm} learning in general. Among the recent efforts, methods drawn inspiration from the diffusion probabilistic models \citep{sohl2015deep,ho2020denoising,song2020improved,song2020score} have demonstrated superior results. In particular, \citet{gao2020learning} propose a diffusion recovery likelihood method to learn and sample from a sequence of \acp{ebm} defined on increasingly noisy versions of a dataset; the models are trained by optimizing conditional likelihoods, which are more tractable than the marginal likelihood. It greatly mitigates the burden of sampling during training. A natural question thus emerges: \textit{Can we leverage the methodology of diffusion models to address the learning issue of energy-based prior?}

In this work, we make the first attempt to address the learning issue of energy-based prior through leveraging diffusion models in the latent space, with a focus on interpretable text modeling. We first unveil the non-trivial symbiosis between latent-space \acp{ebm} and diffusion models. Specifically, we focus on the symbol-vector coupling prior; we construct a flexible process that restores the hidden structure in text data by noise-level-aware sampling from a learned sequence of conditional \acp{ebm} in the latent space. A variational learning framework is then derived from it. We further employ a geometric clustering-based regularization that complements the symbol-inducing information bottleneck to improve the quality of learned latent space. We term the resulting model \ac{ldebm}. Compared to \citet{gao2020learning}, which deals with \acp{ebm} in the data space, \ac{ldebm} is directly applicable to text data with or without labels; it extracts interpretable latent structures that benefit potential downstream tasks such as semi-supervised classification. Although there are methods using diffusion models in the latent space, some of which have achieved very impressive image generation results, \eg, \citet{vahdat2021score}, few of them to our knowledge have explored (unsupervised) symbol induction in the latent space especially on text data. In addition, our method can be trained from scratch and form a well-structured latent space without pretraining, as required by concurrent works on image modeling such as \citet{vahdat2021score} and \citet{nie2021controllable}. 
In our experiments on generative modeling and interpretable text modeling, \ac{ldebm} largely outperforms strong counterparts in terms of both generation quality and interpretability of the learned latent space.

\paragraph{Contributions}

(1) We introduce a novel symbiosis of the latent space \ac{ebm} and diffusion model in a variational learning framework; the model can be trained from scratch, is directly applicable to text data with or without labels, and shows superior sampling quality. (2) We develop a geometric clustering-based regularization jointly with the information bottleneck that tackles the mode-collapse problem in variational learning of the latent space \ac{ebm}. (3) Our experiments demonstrate that the proposed model learns a well-structured latent space and delivers strong results on interpretable text modeling.

\section{Preliminaries: Symbol-Vector Coupling \texorpdfstring{\ac{ebm}}{}}\label{sec:svebm}

We assume that for an observed high-dimensional sample $\x \in \R^D$, there exists $\z \in \mathbb{R}^d$ as its compact continuous latent variables. We assume that $\y$ is the symbolic one-hot vector indicating one of $K$ categories that $\z$ belongs to. The complete-data distribution is
$
    p_\theta(\y, \z, \x) = p_\alpha(\y, \z) p_\beta(\x | \z)
$,
where $p_{\alpha}(\y, \z)$ is the joint prior model with parameters $\alpha$, and $p_\beta(\x|\z)$ is the top-down generation model with parameters $\beta$; henceforth, we use $\theta = (\alpha, \beta)$ to summarize the parameters. Given $\z$, $\y$ and $\x$ are independent; \ie, $\z$ is sufficient for $\y$ in this model. 

\citet{pang2021latent} propose to formulate the joint prior model, $p_{\alpha}(\y, \z)$, as an \ac{ebm},
\begin{equation}
    p_{\alpha}(\y, \z) 
        = \frac{1}{Z_\alpha} \exp(\langle \y, f_\alpha(\z)\rangle) p_0(\z),
    \label{equ:svebm}
\end{equation}
where $p_0(\z)$ is a reference distribution, assumed to be the non-informative prior (\eg, isotropic Gaussian or uniform) of the conventional generation model, $f_{\alpha}(\z) \in \mathbb{R}^K$ is parameterized by a small multi-layer perceptron, and $Z_\alpha$ is the normalizing constant or partition function. The energy term $ \langle \y, f_\alpha(\z)\rangle$ in \cref{equ:svebm} forms an associative memory that couples the symbol $\y$ and the dense vector $\z$. Given $\z$, 
\begin{equation}
    p_\alpha(\y|\z) \propto \exp(\langle \y, f_\alpha(\z)\rangle)
    \label{equ:y_cond_z}
\end{equation}
becomes a softmax classifier, where $f_\alpha(\z)$ provides the logit scores for the $K$ categories. Marginally, we have
\begin{equation}
    p_\alpha(\z) = \frac{1}{Z_\alpha} \exp(F_\alpha(\z))p_0(\z),
    \label{equ:z_ebm}
\end{equation}
where the marginal energy term is in a log-sum-exponential form,
$
    F_\alpha(\z) = \log \sum_\y \exp(\langle \y, f_\alpha(\z)\rangle)
$.
It is shown that the coupling between $\z$ and $\y$ enables a symbol-aware continuous vector computation during prior and posterior sampling, which helps to induce a structural latent space \citep{pang2021latent}. Finally, the prior model $p_\alpha(\y, \z)$ stands on a generation model $p_\beta(\x|\z)$. In text modeling, let $\x = (\x^{(t)}, t=1,...,T)$ be a sentence, where $\x^{(t)}$ is the $t$-th token. $p_{\beta}(\x|\z)$ can be defined as a conditional autoregressive model,
$
    p_\beta(\x|\z) = \prod_{t=1}^T p_\beta(\x^{(t)}|\x^{(1)}, ..., \x^{(t-1)}, \z)
$.
The complete model $p_\theta(\y, \z, \x)$ with the energy-based prior $p_\alpha(\y, \z)$ and the generation model $p_\beta(\x | \z)$ is termed as \ac{svebm}.

In principle, a \ac{svebm} can be learned through maximizing the log-likelihood function, where the learning gradient is
$
    \nabla_\theta \log p_\theta (\x) = \E_{p_\theta(\z|\x)} [
        \nabla_\theta (
            \log p_\alpha (\z) + 
            \log p_\beta (\x|\z)
        )
    ]
$. To estimate the expectation, one may sample from the prior $p_{\alpha}(\z)$ and the posterior $p_{\theta}(\z|\x)$ with Langevin dynamics \citep{welling2011bayesian}. Since $f_\alpha$ is a small network, prior sampling is particularly affordable. In comparison, the posterior sampling can be more expensive as it requires back-propagating through the generation network. One promising solution is to follow the variational learning scheme \citep{kingma2013auto} that amortizes the posterior sampling from $p_\theta(\z|\x)$ by an inference network $q_\phi(\z|\x)$; \ac{mcmc}-based sampling can be used for prior samples.

\section{\texorpdfstring{\acl{ldebm}}{}}

\subsection{A Symbiosis between \texorpdfstring{\ac{svebm}}{} and Diffusion Model}\label{sec:ldebm}

Contrasting to the vanilla sampling process of the latent variables in \ac{svebm}, \ac{ldebm} follows the philosophy of diffusion probabilistic models \citep{sohl2015deep}; it assumes a sequence of perturbed samples, $\z_0, \z_1, ..., \z_T$, to construct a flexible process that restores the structure in data. First, we define the forward diffusion process that systematically and gradually destroys structure in a data distribution: 
$
    \z_0 \sim q_\phi (\z_0|\x); 
    \z_{t+1} = \sqrt{1 - \sigma^2_{t+1}}\z_t + \sigma_{t+1}\bm{\epsilon}_{t+1}
$,
where $t=0,1,...,T - 1$ and $\bm{\epsilon}_t$ is the zero-mean standard Gaussian noise. The scaling factor $\sqrt{1 - \sigma^2_{t+1}}$ ensures that the sequence is a spherical interpolation between the posterior sample and the Gaussian white noise. The forward trajectory and the Markov transition between each perturbed samples $\z_1, ..., \z_T$ are thus
\begin{equation}
    \begin{aligned}
        & q_\phi(\z_{0:T}|\x) 
            = q_\phi(\z_0|\x) \prod\limits_{t=0}^{T-1} q(\z_{t+1}|\z_{t}); \\
        & q(\z_{t+1} | \z_{t}) 
            = \N(\z_{t+1} ; 
                \sqrt{1-\sigma^2_{t+1}}\z_{t}, 
                \sigma^2_{t+1} \I).
    \end{aligned}
    \label{equ:forward}
\end{equation}

Our goal is to learn the generative distribution that describes the same trajectory but in reverse. Inspired by \citet{gao2020learning}, we start by constructing a sequence of \textit{marginal} \acp{ebm} at each diffusion step in the latent space. The \textit{conditional} \acp{ebm} aiming at recovering $\z_0$ from noisy inputs then follow as (see the derivation in \cref{app:cond_ebm}):
\begin{equation}
    \begin{aligned}
        & p_\alpha( \tilde{\z}_t|\z_{t+1} ) = \\ 
                & \frac{1}{\Tilde{Z}_{\alpha, t}(\z_{t+1})} 
                \exp{
                \left(
                    F_\alpha(\tilde{\z}_t, t) - 
                    \frac{1}{2\sigma^2_{t+1}}||\tilde{\z}_t - \z_{t+1}||^2
                \right)},
    \end{aligned}
    \label{equ:cond_ebm}
\end{equation}
where $t=0,1,...,T-2$. We denote
$
    \tilde{\z}_t = 
    \sqrt{1 - \sigma^2_{t+1}}\z_t
$  for brevity.
$
    F_\alpha(\tilde{\z}_t, t)
$ is the neural network that parameterizes the energy function at each diffusion step, and 
$
    \Tilde{Z}_{\alpha, t}(\z_{t+1}) = 
        \int 
            \exp{(
                F_\alpha(\tilde{\z}_t, t) - \frac{1}{2\sigma^2_{t+1}} ||\tilde{\z}_t - \z_{t+1}||^2
            )} d\tilde{\z}_t
$ 
is the partition function of each conditional \ac{ebm}. For $t=T-1$,
$
    p_\alpha( \tilde{\z}_{t}|\z_{t+1} ) = 
        \frac{1}{\Tilde{Z}_{\alpha, t}} 
            \exp{(
                F_\alpha(\tilde{\z}_{t}, t) - 
                \frac{1}{2\sigma^2_{t+1}}||\tilde{\z}_{t}||^2
            )}
$
since the diffused samples at time step $T$ should be close to Gaussian white noise; the distribution of $\tilde{\z}_{T-1}$ can thus be exponentially tilting of a zero-mean Gaussian distribution.

\cref{equ:cond_ebm} shares the idea of denoising generative modeling \citep{bengio2013generalized}, where a denoising autoencoder is trained by maximizing the conditional probabilities of the observed samples given their noisy versions. Compared to the vanilla definition (see \cref{equ:z_ebm}), the noise-level-aware quadratic term constrains the energy landscape to be localized around the noisy sample; this makes the latent space much less multi-modal and easier to sample from. To be specific, \citet{gao2020learning} show that $p_\alpha( \tilde{\z}_t|\z_{t+1} )$ is approximately a single-mode Gaussian distribution when $\sigma$ is sufficiently small; it greatly reduces the burden of \ac{mcmc} sampling. After sampling $\tilde{\z}_t$ from the model, we can easily obtain $\z_t = \tilde{\z}_t / \sqrt{1 - \sigma^2_{t+1}}$.

Next, we show that the forward and reverse process in the latent space can be naturally integrated into the variational learning scheme to amortize the time-consuming posterior sampling. Similar to \ac{vae}, the ELBO in \ac{svebm} is
\begin{equation}
    \begin{aligned}
    &\ELBO_{\theta, \phi} 
          = \log p_\theta(\x) - \KL( q_\phi(\z|\x) \| p_\theta(\z|\x) ) \\
         &= \E_{q_\phi(\z|\x)} [ \log p_\beta(\x|\z) ] 
          - \KL( q_\phi(\z|\x) \| p_\alpha(\z) ),
    \end{aligned}
    \label{equ:elbo_svebm}
\end{equation}
where $\KL$ denotes the Kullback-Leibler divergence. Since we now consider the full trajectory of the perturbed samples, in \ac{ldebm} we may optimize
\begin{equation}
    \resizebox{0.9\hsize}{!}{$%
        \begin{aligned}
            \displaystyle
            \ELBO_{{\rm Diff}, \theta, \phi} &= 
                  \E_{q_\phi (\z_0|\x)} \left[ 
                        \log p_\beta(\x|\z_0) - 
                        \log q_\phi (\z_0|\x) \right] \\
                    & + \E_{q_\phi (\z_0|\x), q(\z_{1:T}|\z_0)} 
                        \left[ 
                            \log \frac{ p_\alpha(\z_{0:T})}{q (\z_{1:T}|\z_0)} 
                        \right],
        \end{aligned}%
    $}%
    \label{equ:elbo_ldebm}
\end{equation}
which is a valid ELBO by applying Jensen's inequality to \cref{equ:elbo_svebm}. The joint training of inference, prior and generation models can be largely reduced to finding the agreement of the forward and reverse Markov transitions defined by $q_\phi$ and $p_\theta$, respectively. Please refer to \cref{app:elbo} for more detailed derivations and discussions.

Finally, we show how to introduce the symbolic one-hot vector $\y$ into our formulation. We assume a complete data distribution that considers the full trajectory of the perturbed latent variables, $p_\theta(\y, \z_{0:T}, \x)$. Among several possibilities for coupling the symbolic vector $y$ with the latent variables, two major options arise: We can couple the symbol with the whole trajectory, \ie, $p_\theta(\y, \z_{0:T}, \x) = p_\alpha(\y, \z_{0:T}) p_\beta(\x | \z_{0:T})$; or we can couple the symbol with only the clean posterior sample $\z_0$, \ie, $p_\theta(\y, \z_{0:T}, \x) = p(\z_T) p_\alpha(\y, \z_{0}|\z_1)\prod_{t=1}^{T-1}p_\alpha(\z_{t}|\z_{t+1}) p_\beta(\x | \z_{0})$. We prefer the latter one, since it is sufficient to model the reverse Markovian transition, while enabling a simpler and more efficient training scheme following \citet{ho2020denoising} (see \cref{sec:alg}). Of note, coupling only $\z_0$ to $\y$ means that we condition only the final reverse diffusion step $[\z_0|\z_1]$ on $\y$ when performing controllable generation. This could be a bit counter-intuitive as no label information is injected in previous reverse steps. Theoretically, $\y$ and $\z_{1:T}$ are independent given $\z_0$ in our formulation; however, we empirically observe that $\y$ and $\z_t$ for $t > 0$ are nearly independent even marginally, after we integrating out $\z_{0:t-1}$ in our model. In other words, $p_\alpha(\y|\z_t),~t>0$ are in general non-informative since adding noise in the latent space could be much more corrupting than adding noise in the data space. The model learns to enjoy the less multi-modal energy landscape in previous reverse steps; it then seeks the given mode only in the most informative final reverse step. Specifically, we achieve this coupling by similarly defining $p_\alpha(\y,\z_0|\z_1)$ as in \cref{equ:svebm} and using the log-sum-exponential form for learning as in \cref{equ:z_ebm}. Please refer to \cref{fig:pgm} for a graphical illustration of our model and \cref{app:symb_coup,app:extra_details} for more details and discussions.

\subsection{\texorpdfstring{\acl{ib}}{}}\label{sec:ib}

To learn the symbolic vector $\y$, we may consider adopting the \acf{ib} principle \citep{tishby2000information}, an appealing approach for inducing symbolic representations. In this section, we re-interpret the above ELBO as a cooperative learning objective, defined as the divergence between two joint distributions; we then show how this formulation helps to incorporate the \ac{ib}-based regularization into \ac{ldebm} in a principled manner.

As shown in \citet{han2019divergence}, the variational learning scheme can be regarded as performing alternating projection between two joint distributions, $Q_\phi$ and $P_\theta$. In our modeling, we have: $Q_\phi(\x, \z_{0:T}) = q_{\rm data}(\x) q_\phi(\z_{0:T} | x)$, and $P_\theta(\x, \z_{0:T}) = p(\z_T) \prod_{t=0}^{T-1}p_\alpha(\z_{t}|\z_{t+1}) p_\beta(\x | \z_{0})$; we use $q_{\rm data}(\x)$ to denote the data distribution of $\x$ for notation consistency. Maximizing $\E_{q_{\rm data}(\x)}[\ELBO_{{\rm Diff}, \theta, \phi}(\x)]$ over $(\theta, \phi)$ is equivalent to minimizing the following divergence:
\begin{equation}
    \begin{aligned}
    \KL( & Q_\phi \| P_\theta )
         = \KL(q_{\rm data}(\x) \| p_\theta(\x)) \\
        &+ \E_{q_{\rm data}(\x)} [\KL(q_\phi(\z_{0:T} | \x) \| p_\theta(\z_{0:T} | \x))],
    \end{aligned}
    \label{equ:kld_joint}
\end{equation}
since $\HH(\x) = - \E_{q_{\rm data}(\x)} [\log q_{\rm data}(\x)]$, \ie, the entropy of data distribution is fixed. Minimizing the KL-divergence $\min_\theta \min_\phi \KL (Q_\phi\|P_\theta)$ defines a cooperative game, with the dynamics that $q_\phi$ and $p_\theta$ run towards each other. 

Since the initial posterior sample $\z_0$ is coupled with the symbolic vector $\y$, it should be the most informative latent variable for inducing the discrete symbol. We can therefore plug in \cref{equ:kld_joint} with a mutual information term between $\z_0$ and $\y$: $\II(\z_0, \y) = \HH(\y) - \HH(\y|\z_0)$, which essentially incorporates the \ac{ib} as we show below. Given the distribution $Q_\phi(\x, \z_{0:T})$, we can first define the marginal distribution of $\z_0$ as the aggregated posterior by integrating out $\z_{1:T}$: 
$
    q_\phi(\z_0) = \E_{q_{\rm data}(\x)} [q_\phi(\z_0 | \x)] 
$.
The entropy of $\z_0$ and conditional entropy of $\z_0$ on $\x$ then follow as $\HH(\z_0)$ and $\HH(\z_0 | \x)$, respectively. Taken together, the KL-Divergence with $\lambda\II(\z_0, \y)$ can therefore be parsed as
\begin{equation}
    \begin{aligned}  
        \mathcal{L} &= \KL(Q_\phi \| P_\theta) - \lambda \II(\z_0, \y) \\
          &= \mathcal{C} + \mathcal{L}_\text{RC}
          + \mathcal{L}_\text{EBM} 
          + \mathcal{L}_\text{IB},
    \end{aligned}
    \label{equ:kld_mi}
\end{equation}
where 
$
    \mathcal{C} = - \HH(\x) +\sum_{t=0}^{T-1} \HH(\z_{t+1}|\z_{t})
$ does not involve learnable parameters,
$
    \mathcal{L}_\text{RC} = -\E_{Q_\phi} [\log p_\beta(\x|\z_{0})]
$ is the reconstruction loss,
$
    \mathcal{L}_\text{EBM} = \KL(q_\phi(\z_{0}) \| p_\alpha(\z_{0:T})) 
$ corresponds with learning latent space models, and
$
    \mathcal{L}_\text{IB} = \II(\x, \z_0) - \lambda\II(\z_0, \y)
$ is the \ac{ib}, where
$
    \II(\x, \z_0) = \HH(\z_0) - \HH(\z_0 | \x)
$
is the mutual information between $\x$ and $\z_0$ under $Q_\phi$; $\lambda \ge 0$ controls the expressivity of $\z_0$ to $\y$. Please refer to \cref{app:ib} for more details. 

\subsection{Geometric Clustering Anchors the Modes}\label{sec:gc}

As shown in the previous section, \ac{ib} provides an elegant solution for inducing the symbolic vector $\y$. In this section, we further introduce an approach that facilitates the emergence of $\y$ from a geometric perspective. To induce a latent space with interpretable structures, ideally, the location of data points in the latent space encodes their semantic meaning, \ie, it indicates the semantic class; semantically similar points should be placed closer and produce the same symbolic vector $\y$. This resembles geometric clustering algorithms: Labels of data points are assigned based on their geometric (typically Euclidean) distance from each other. Below, we show how to realize this intuition in \ac{ldebm}.

Let us consider the joint distribution $p_\theta(\x, \y)$. We can decompose its log-likelihood into 
$
    \log p_\theta(\x, \y) = \log p_\theta(\x) + \log p_\theta(\y|\x)
$ as in \citet{grathwohl2019your}, where $\log p_\theta(\x)$ is substituted with the ELBO derived in \cref{sec:ldebm}. $p_\theta(\y|\x)$ is the classification model in the latent space:
$
    p_\theta(\y|\x) \approx \E_{q_\phi(\z_0|\x)} [p_\alpha(\y|\z_0)]
$.
$p_\alpha(\y|\z_0)$ is the softmax classifier of $\y$ based on $\z_0$ similarly as in \cref{equ:y_cond_z}, detailed in \cref{app:symb_coup}. Therefore, we can encode the semantic information from the label $\y$ into $\z_0$ through learning the classifier $p_\alpha(\y|\z_0)$. In case there is full or partial access to the ground-truth semantic class labels, we could directly utilize these labels to supervise the classifier, jointly with the existing ELBO objective. When no label is provided, we generate pseudo label $\hat{\y}$ by clustering $\z_0$, which optimizes $\E_\y \log p_\theta(\x, \y)$ instead; $\E_\y$ is defined by the clustering algorithm. It is akin to the EM algorithm, where geometric clustering serves as a hard-decision E-step to help induce $\y$. In practice, we employ K-means to cluster $\z_0$. In \cref{sec:gen_model}, we empirically show that this strategy learns a better latent space and significantly alleviates the mode-collapse problem.

\subsection{Algorithms and Implementation}\label{sec:alg}

\begin{algorithm}[ht!]
    \small
    \caption{\textbf{Learning algorithm.}}
    \label{alg:learn}
    \begin{algorithmic}
    \STATE {\bfseries input:} 
        initial parameters $(\alpha, \beta, \phi)$, 
        learning rate $\eta$, 
        observed unlabeled examples $\{\x^{(i)}\}_{i=1}^M$, 
        observed labeled examples $\{(\x^{(i)}, \y^{(i)})\}_{i=M+1}^{M+N}$ 
        (alternative, needed in controllable generation or semi-supervised learning).
    \REPEAT
    \STATE {\bfseries posterior sampling:} 
        For each $\x^{(i)}$, sample $\z^{(i)}_0 \sim q_\phi(\z_0|\x^{(i)})$ using inference network.
    \STATE {\bfseries prior sampling:} 
        For each $\z^{(i)}_0$, sample diffusion step $t$ from $\text{Unif}(\{0, ..., T - 1\})$, and the perturbed pair $(\tilde{\z}^{(i)}_t, \z^{(i)}_{t+1})$ following \cref{equ:forward}. Set $\tilde{\z}^{(i)}_t$ as the positive sample $\tilde{\z}^{(i)+}_t$. Initialize the \ac{mcmc} using $\z^{(i)}_{t+1}$ and update by \cref{equ:prior_lang} for $K$ steps to obtain $\tilde{\z}^{(i)-}_{t}$.
    \STATE {\bfseries learning prior model:} Update $\alpha$ with \\
        $
            \eta (\sum_{i} [
                \nabla_\alpha F_\alpha(\tilde{\z}^{(i)+}_{t}, t) 
              - \nabla_\alpha F_\alpha(\tilde{\z}^{(i)-}_{t}, t) 
            ] - \nabla_\alpha \II)
        $. 
    \STATE {\bfseries learning inference and generation models:} \\
        Update $\beta$ and $\phi$ with \cref{equ:poste_grad} and $\nabla_{\phi} \II$.
    
    \IF{labeled data $(\x^{(i)}, \y^{(i)})$ is available}
        \STATE {\bfseries update $\gamma = (\alpha, \phi)$ using $\y^{(i)}$:} \\ 
        Learning gradient 
        $
            \eta \sum_{i} \nabla_\gamma \log p_{\alpha_t}(\y^{(i)}|\z^{(i)}_0)
        $ is provided by ground-truth label.
    \ELSIF{only unlabeled data is available}
        \STATE {\bfseries update $\gamma = (\alpha, \phi)$ 
                using pseudo-label $\hat{\y}^{(i)}$:} \\
        Geometric clustering generates $\hat{\y}^{(i)}$ for each $\x^{(i)}$.
        $
            \eta \sum_{i} \nabla_\gamma \log p_{\alpha_t}(\hat{\y}^{(i)}|\z^{(i)}_0)
        $, \ie, the gradient comes from pseudo-label generated by geometric clustering.
    \ENDIF
    \UNTIL{converged.}
    \end{algorithmic}
\end{algorithm}

\paragraph{Training and sampling algorithms}

For learning the prior model, we have for each $t=0,1,...,T - 1$:
\begin{equation}
        \begin{aligned}
            \displaystyle
            \nabla_\alpha \ELBO_t &= 
                 \E_{q_\phi(\tilde{\z}_t, \z_0|\x)} 
                    [\nabla_\alpha F_\alpha(\tilde{\z}_t, t)] \\
              &- \E_{q_\phi(\z_{t+1}, \z_0|\x), p_\alpha(\tilde{\z}_t|\z_{t+1})}
                    [\nabla_\alpha F_\alpha(\tilde{\z}_t, t)].
        \end{aligned}%
    \label{equ:prior_grad}
\end{equation}

Let $\psi = \{\beta, \phi\}$ collect the parameters of the inference (encoder) and generation (decoder) models.
\begin{equation}
        \begin{aligned}
            \displaystyle
            &\nabla_\psi \ELBO 
                = \nabla_\psi \E_{q_\phi(\z_0|\x)} [
                    \log p_\beta(\x|\z_0) - \log q_\phi (\z_0|\x)
                ] \\
            &   -\nabla_\phi \E_{q_\phi (\z_{0:T}|\x)} 
                     \left[ 
                        \log p (\z_T) + 
                        \sum_{t=0}^{T-1} \log p_\alpha (\z_{t}|\z_{t+1})
                     \right].
        \end{aligned}%
    \label{equ:poste_grad}
\end{equation}

Recall that we denote $\tilde{\z}_t = \sqrt{1 - \sigma^2_{t+1}}\z_t$. $\E_{p_\alpha(\tilde{\z}_t|\z_{t+1})}$ is approximated by \ac{mcmc} samples from the prior.  $\E_{q_\phi(\z_0|\x)}$ is approximated by samples from the inference network. We also add the gradient from $\II(\z_0, \y)$, denoted as $\nabla \II$, to \cref{equ:prior_grad,equ:poste_grad} during training to incorporate \ac{ib}. Please see \cref{app:grad} for detailed derivations.

Note that the expectation in \cref{equ:prior_grad} requires \ac{mcmc} sampling (\eg, Langevin dynamics \citep{welling2011bayesian}) of the prior model. For a target distribution $\pi(\tilde{\z})$, the dynamics iterates 
$
    \tilde{\z}^{k+1} = \tilde{\z}^k 
                    + \frac{s^2}{2} \nabla_{\tilde{\z}} \log \pi (\tilde{\z}^k)
                    + s \bm{\epsilon}^k
$, 
where $k$ indexes the iteration of the dynamics, $s$ is a small step size, and $\bm{\epsilon}^k \sim \N(0, \I)$ is the Gaussian noise. In this work, we follow the heuristics in \citet{gao2020learning} and set the step size $s_t = b\sigma_t c_t$, where $b < 1$ is a tuned hyperparameter, and $c_t = \sqrt{\prod_{i=1}^t \sigma_i / \sigma_1}$ is a scaling factor. Let $t$ indexes the diffusion step; $K$ steps of Langevin dynamics thus iterates
\begin{equation}
    \resizebox{0.88\hsize}{!}{$%
        \begin{aligned}
            \displaystyle
            \tilde{\z}_t^{k + 1} &= \tilde{\z}_t^k
                    + \frac{b^2\sigma_t^2 c_t^2}{2}\left(
                        \nabla_{\tilde{\z}} F_\alpha(\tilde{\z}_t^k, t) - \frac{1}{\sigma_t^2}(\tilde{\z}_t^k - \z_{t+1})
                      \right) \\
                  &+ b\sigma_t c_t \bm{\epsilon}^k.
        \end{aligned}%
    $}
    \label{equ:prior_lang}
\end{equation}

\begin{algorithm}[ht!]
    \small
    \caption{\textbf{Synthesizing algorithm.}}
    \label{alg:synth}
    \begin{algorithmic}
       \STATE {\bfseries input:} $\z_T \sim \N(0, \I)$
       \STATE {\bfseries output:} $\z_0$
       \FOR{$t=T-1$ {\bfseries to} $t=0$}
           \STATE Initialize $\tilde{\z}_t = \z_{t+1}$.
               \FOR{$k=1$ {\bfseries to} $K$}
                   \STATE Update $\tilde{\z}_t$ using \cref{equ:prior_lang}.
               \ENDFOR
           \STATE $\z_t = \tilde{\z}_t / \sqrt{1 - \sigma^2_{t+1}}$
       \ENDFOR
    \end{algorithmic}
\end{algorithm}

In principle, training the model amounts to minimizing the ELBO in \cref{equ:elbo_ldebm}, which requires a summation over all the diffusion steps; it involves sampling a full forward trajectory. To make the training simpler and more efficient, following \citet{ho2020denoising}, we randomly choose one diffusion step from the summation to optimize at each training iteration. After training, we initialize the reverse trajectory from Gaussian white noise. The synthesized sample at each step serves to initialize an \ac{mcmc} that samples from the model of the previous step. The learning and synthesizing algorithms are summarized in \cref{alg:learn,alg:synth}, respectively.

\paragraph{Implementation}

For the K-means algorithm, we use the implementation of \citet{johnson2019billion}, which explicitly deals with the empty clusters and trivial parameterization problems. To emphasize that the proposed model shows better capability of modeling latent space, we use the same encoder and decoder as \citet{pang2021latent} for all the experiments. We use a shared network $F_\alpha(\tilde{\z}_t, t)$ for each $t=0,1,...,T - 1$; $T=6$; $t$ is encoded by sinusoidal position embedding as in \citet{ho2020denoising}, and we set $\sigma_t^2$ to increase linearly. For Langevin dynamics, we use $K=50$ and $b^2=0.002$ throughout the experiments. See \cref{app:impl} for network architecture and further training details.

\section{Experiments}

Through a series of experiments, we empirically examine the capability of our model for generative modeling and interpretability on text modeling tasks. Please refer to \cref{app:db} for additional experiment settings and baselines.

\subsection{Generative Modeling}\label{sec:gen_model}

\paragraph{2D synthetic data}

\begin{figure}[b!]
    \centering
    \includegraphics[width=0.9\linewidth]{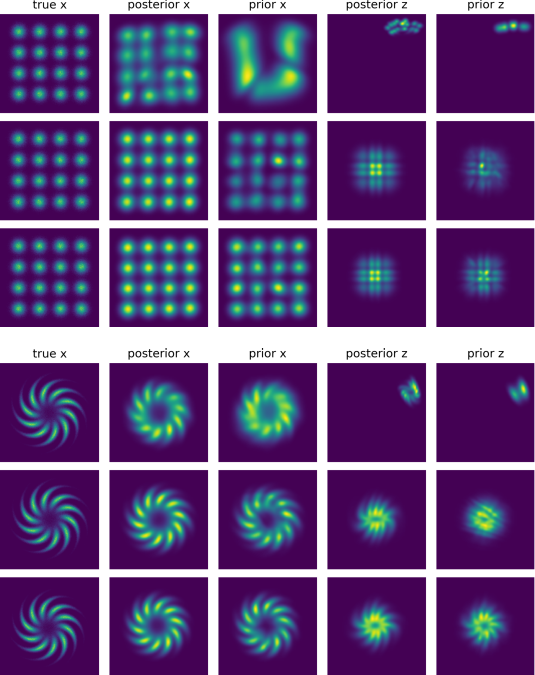}
    \caption{\textbf{Evaluation on 2D synthetic data}: a mixture of 16 Gaussians (upper panel) and a 10-arm pinwheel-shaped distribution (lower panel). In each panel, the top, middle, and bottom row display densities learned by \ac{svebm}-\ac{ib}, our model w/o geometric clustering, and our full model, respectively. In each row, from left to right, it displays the data distribution and the \acp{kde} of: $\x$ generated by amortized posterior $\z$ samples, $\x$ by \ac{mcmc} sampled prior $\z$ samples, posterior $\z$ samples, and prior $\z$ samples.}
    \label{fig:toy_data}
\end{figure}

\begin{figure}[ht!]
    \centering
    \includegraphics[width=0.9\linewidth]{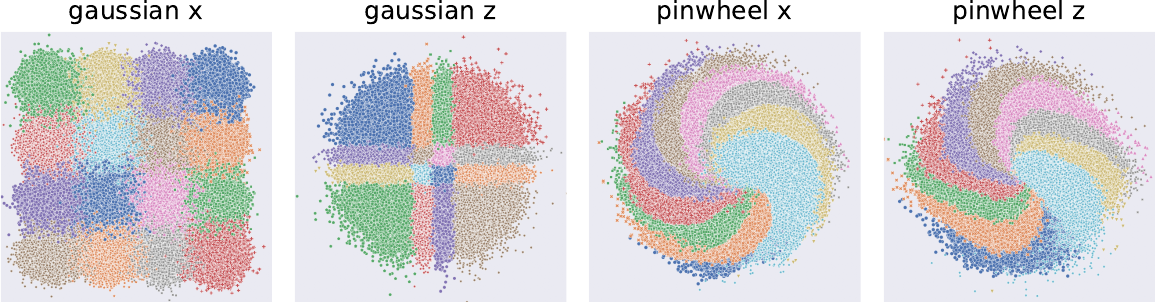}
    \caption{\textbf{Visualization of color-coded data points.} We visualize data points and the corresponding inferred latent variables of two 2D synthetic datasets (\textit{gaussian} and \textit{pinwheel}). Data points with different labels are assigned with different colors.}
    \label{fig:labeled_data}
\end{figure}

We first perform experiments of our model on 2D synthetic datasets as a sanity check to validate our assumptions; results are displayed in \cref{fig:toy_data}. The gap between \ac{ldebm} and \ac{svebm} is very clear. As mentioned in \cref{sec:intro}, for more complex datasets (\eg, datasets with more modes or more complex data structure), \ac{svebm} struggles to capture regularities in the data; the model is prone to collapse, which features an exploding KL-term and poor performance on generation. We provide more results that show the full evolution of these models during training with more discussions in \cref{app:extra_details}. In contrast, \ac{ldebm} without geometric clustering already overcomes this problem, performing relatively well in terms of modeling both \textit{posterior} $\x$ and \textit{prior} $\x$. Although \ac{ldebm} without geometric clustering faithfully reconstructs the data and shows significant improvement on generation quality, the generated distribution can be slightly distorted, and some modes are missing. The problem is clearer in the latent space: Mode-collapse occurs in the \textit{prior} $\z$ distribution, where the latent structure is broken. \ac{ldebm} with geometric clustering maintains the number of modes as in the data distribution and induces a highly-structural latent space, echoing our intuition in \cref{sec:gc}. \cref{fig:labeled_data} shows the structural similarity between data distribution and the learned latent distribution. 

\paragraph{Language generation}

Following previous state-of-the-art competitors \citep{zhao2018unsupervised,shi2020dispersed,pang2021latent}, we evaluate the quality of generation on a real-world text dataset, \ac{ptb} \citep{marcus1993building} as pre-processed by \citet{mikolov2010recurrent}. We report four metrics of the generation performance: \ac{rppl} \citep{zhao2018adversarially}, BELU \citep{papineni2002bleu}, \ac{wkl}, and \ac{nll}; \cref{tab:ptb_gen} summarizes results.

\begin{table}[ht!]
    \caption{\textbf{Results of language generation on \acs{ptb} dataset.} We highlight our model results in {\color{tgray} gray} color. The best and second-best performances are marked in bold numbers and underlines, respectively; tables henceforth follows this format.}
    \vskip 0.1in
    \label{tab:ptb_gen}
    \centering
    \small
    \begin{sc}
        \begin{tabular}{lcccc}
            \toprule
            { Model} & {\bf rPPL$^\downarrow$} & {\bf BLEU$^\uparrow$} & {\bf wKL$^\downarrow$} & {\bf NLL$^\downarrow$} \\
            \midrule
            { Test Set} & - & 100.0 & 0.14 & - \\ 
            \midrule
            { RNN-LM} & -  & - & - & 101.21 \\
            \midrule 
            { AE} & 730.81 & 10.88 & 0.58 & - \\ 
            { VAE} & 686.18 & 3.12 & 0.50 & 100.85 \\ 
            \midrule
            { DAE } & 797.17 & 3.93 & 0.58 & - \\ 
            { DVAE} & 744.07 & 1.56 & 0.55  & 101.07 \\ 
            { DI-VAE} & 310.29 & 4.53 & 0.24 & 108.90 \\
            \midrule
            { semi-VAE} & 494.52 & 2.71 & 0.43 & 100.67 \\ 
            { semi-VAE $+ \II$} & 260.28 & 5.08 & 0.20 & 107.30 \\
            { GM-VAE} & 983.50 & 2.34 & 0.72 & 99.44 \\
            { GM-VAE $+ \II$} & 287.07 & 6.26 & 0.25 & 103.16 \\
            { DGM-VAE } & 257.68 & 8.17 & 0.19  & 104.26 \\
            { DGM-VAE $+ \II$} & 247.37 & 8.67 & 0.18 & 105.73 \\ 
            { \ac{svebm} } & 180.71 & 9.54 & 0.17 & 95.02 \\
            { \ac{svebm}-IB } & 177.59 & 9.47 & 0.16 & 94.68 \\
            \midrule
            \rowcolor{gray}
            { Ours w/o GC } & \underline{168.32} & \underline{11.12} & \underline{0.07} & \textbf{79.84} \\
            \rowcolor{gray}
            { Ours } & \textbf{164.57} & \textbf{11.16} & \textbf{0.06} & \underline{82.38} \\
            \bottomrule
        \end{tabular}
    \end{sc}
\end{table}

The proposed model, either w/ or w/o geometric clustering, demonstrates the best performance on reconstruction (highest BLEU) and fitting capacity (lowest \ac{nll}) than all baseline models. Moreover, the higher expressivity of our models enables the generation of high-quality sentences. The lowest \ac{rppl} indicates that our models can further improve over these strong baselines on fluency and diversity of generated text; the lowest \ac{wkl} indicates that the word distribution of the generated sentences is the most consistent with that of the original data.

\paragraph{Sentence completion}

Further, we show that the trained model enables text completion on a masked JerichoWorld dataset \citep{ammanabrolu2021modeling}. We perform conditional sampling in the latent space to complete the masked sentences; please see more details in \cref{app:extra_details} and \cref{tab:jericho_sent_completion}.

\begin{table}[ht!]
    \caption{\textbf{Sentence completion on JerichoWorld dataset.} The {\color{tgray} gray} words in the input sentences are masked with \texttt{<unk>} token.}
    \label{tab:jericho_sent_completion}
    \vskip 0.1in
    \centering
    \small
    \begin{tabular}{ll}
        \toprule
        \multirow{4}{*}{\bf Input} 
        & ... A bathroom lies to the south, while a door\\
        & to the east leads to {\color{tgray} the living room. On the bed} \\
        & {\color{tgray} are a driver's license, some keys and a wallet} \\
        & {\color{tgray} On the end table is a telephone.} \\
        \midrule
        \multirow{4}{*}{\bf Pred.} 
        & ... A bathroom lies to the south, while a door \\
        & to the east leads to the living room. On the bed \\ 
        & is a wallet. On the end table are a telephone \\ 
        & and some keys. \\
        \bottomrule
        \toprule
        \multirow{3}{*}{\bf Input} 
        & ... All around you the crowd is in a state of \\
        & {\color{tgray} pandemonium. The paths of least resistance} \\
        & {\color{tgray} are up, down and west.} \\
        \midrule
        \multirow{3}{*}{\bf Pred.} 
        & ... All around you the crowd is in a state of\\
        & pandemonium. The paths of least resistance \\
        & are down and east. \\
        \bottomrule
    \end{tabular}
\end{table}

\subsection{Interpretable Text Modeling}\label{sec:inter_model}

In this section, we move on to evaluate our model on the interpretability of text modeling.

\paragraph{Unsupervised text attributes discovery}

First, we examine the efficacy of our model on the unsupervised text attributes discovery task. We assess the model on the \ac{dd} dataset \citep{li2017dailydialog}, a chat-oriented dataset of 13,118 daily conversations with human-annotated dialog action and emotion labels for the utterances. The interpretability is evaluated through the ability to unsupervisedly capture the utterance attributes of \ac{dd}. We flatten the dialogues for text modeling and use $p_\theta(\y|\x)$ to infer the utterance label. In particular, we take the ${\rm argmax}$ of the classification head as the inferred label. Following \citet{zhao2018unsupervised}, we recruit homogeneity to evaluate the consistency between ground-truth action and emotion labels and those inferred from our model. \cref{tab:dd_cluster} displays the results of our model and baselines. It shows that the proposed model outperform other baselines in reconstruction by a large margin and give a much better homogeneity on both the dialog action and emotion. The superior performance of \ac{ldebm} equipped with latent space geometric clustering again verifies our intuition in \cref{sec:gc}.

\begin{table}[ht!]
    \caption{\textbf{Results of interpretable text modeling on \ac{dd}.} We use mutual information (MI), BLEU, and homogeneity with actions and emotions for evaluation.}
    \label{tab:dd_cluster}
    \vskip 0.1in
    \centering
    \begin{small}
    \small
    \begin{sc}
        \begin{tabular}{lcccc}
            \toprule
            Model  & {\bf MI$^\uparrow$} & {\bf BLEU$^\uparrow$} & {\bf Act.$^\uparrow$} & {\bf Emo.$^\uparrow$} \\
            \midrule
            { \text{DI-VAE}} &   1.20 & 3.05 & 0.18 & 0.09 \\
            \midrule
            { semi-VAE} &  0.03 & 4.06 & 0.02 & 0.08 \\ 
            { semi-VAE $+ \II$} &   1.21 & 3.69 & 0.21 & 0.14 \\ 
            { GM-VAE} &   0.00 & 2.03 & 0.08 & 0.02 \\
            { GM-VAE $+ \II$}  &  1.41 & 2.96 & 0.19 & 0.09 \\
            { DGM-VAE } &   0.53 & 7.63 & 0.11 & 0.09 \\ 
            { DGM-VAE $+ \II$} & 1.32 & 7.39 &  0.23  & 0.16 \\
            { \ac{svebm}} & 0.01 & 11.16 &  0.03  &  0.01 \\
            { \ac{svebm}-IB} & 2.42 & 10.04 & 0.59  & 0.56 \\
            \midrule
            \rowcolor{gray}
            { Ours w/o GC } & \underline{2.44} & \underline{16.72} & \underline{0.65} & \underline{0.63} \\
            \rowcolor{gray}
            { Ours} & \textbf{3.94} & \textbf{28.75} & \textbf{0.74} & \textbf{0.74} \\
            \bottomrule
        \end{tabular}
    \end{sc}
    \end{small}
\end{table}

\paragraph{Conditional response generation} 

Next, we evaluate our model on dialog generation with \ac{smd} \citep{eric2017key} and \ac{dd} datasets. We evaluate the quality of generated responses using BELU and three word-embedding-based topic similarity metrics \citep{shi2020dispersed}: embedding average \citep{mitchell2008vector}, embedding extrema \citep{forgues2014bootstrapping}, and embedding greedy \citep{rus2012optimal}. \cref{tab:smdd_cond_gen} shows that \ac{ldebm} has competitive performance compared with \ac{svebm}-\ac{ib} on \ac{smd} and outperforms the strong baselines on all metrics on \ac{dd}; see qualitative examples in \cref{tab:smd_action_cases,tab:smdd_response_cases}.

\begin{table}[ht!]
    \caption{\textbf{Dialog evaluation results on \ac{smd} and \ac{dd}.} Models are assessed using four metrics: BLEU, average, extrema, and greedy word embedding based similarity.}
    \label{tab:smdd_cond_gen}
    \vskip 0.1in
    \centering
    \small
    \begin{sc}
        \begin{tabular}{cLCCCC}
            \toprule
            \rowcolor{white}
            Data & Model & {\bf BLEU$^\uparrow$} & {\bf Avg.$^\uparrow$} 
                         & {\bf Extr.$^\uparrow$} & {\bf Grdy.$^\uparrow$} \\
            \midrule
            \rowcolor{white}
            & DI-VAE & 7.06 & 76.17 & 43.98 & 60.92 \\
            \rowcolor{white}
            & DGM $+ \II$ & 10.16 & 78.93 & 48.14 & 64.87 \\
            \rowcolor{white}
            & SVE-IB & \textbf{12.01} & \textbf{80.88} & \underline{51.35} & \underline{67.12} \\
            \cmidrule(l){2-6}
            & w/o GC & 11.44 & \underline{80.16} & 51.26 & 66.51 \\
            \multirow{-6}{*}{\ac{smd}} & Ours & \underline{11.51} & \textbf{80.88} & \textbf{51.57} & \textbf{67.13} \\
            \midrule
            \rowcolor{white}
            & DGM $+ \II$ & 2.19 & 74.73 & \underline{45.85} & \underline{64.28} \\
            \rowcolor{white}
            & SVE-IB & \underline{2.23} & \underline{77.37} & 43.32 & 63.99 \\
            \cmidrule(l){2-6}
            \multirow{-3}{*}{\ac{dd}} & Ours & \textbf{3.72} & \textbf{78.89} & \textbf{46.19} & \textbf{65.87} \\
            \bottomrule
        \end{tabular}
    \end{sc}
\end{table}

\begin{table}[ht!]
    \caption{\textbf{Samples of unsupervisedly discovered action categories and corresponding utterances on \ac{smd}.}}
    \label{tab:smd_action_cases}
    \vskip 0.1in
    \centering
    \small
    \begin{tabular}{ll}
        \toprule
        {\bf Action} & Request-weather \\
        \midrule
        \multirow{7}{*}{\bf Utterance} 
        & I need to know if it is going to be foggy \\ 
        & in Fresno today and tomorrow car. \\
        \cmidrule(l){2-2}
        & Manhattan, please. \\ 
        & Will it be cloudy on Monday?  \\
        \cmidrule(l){2-2}
        & I need current weather data about \\
        & New York, specifically information \\ 
        & about the temperature. \\
        \bottomrule
        \toprule
        {\bf Action} & Request-city \\
        \midrule
        \multirow{5}{*}{\bf Utterance} 
        & In what city are you interested? \\
        \cmidrule(l){2-2}
        & What city would you like to know \\ 
        & the weather about? \\
        \cmidrule(l){2-2}
        & Okay, what city should I look in? \\
        \bottomrule
    \end{tabular}
\end{table}

\begin{table}[ht!]
    \caption{\textbf{Dialog cases generated by \ac{ldebm} given the context.} On \ac{smd}, we provide the same context but with different $\y$ values to generate each response; actions indicated by $\y$ are listed in parentheses. On \ac{dd}, \ac{ldebm} can well capture the dialog topic; we provide the ground-truth response in each case for reference.}
    \label{tab:smdd_response_cases} 
    \vskip 0.1in
    \centering
    \small
    \begin{tabular}{ll}
        \toprule
        \multicolumn{2}{l}{\ac{smd}} \\
        \midrule
        \multirow{2}{*}{\bf Ctx.} 
        & \textit{User:} What gas stations are here? \\
        & \textit{Sys:} There is a Chevron. \\
        \midrule
        \multirow{2}{*}{\bf Ref.} 
        & That's good! Please pick the quickest \\
        & route to get there and avoid all heavy traffic! \\
        \midrule
        \multirow{2}{*}{\bf Pred.} 
        & (Req.-address) What is the address? \\
        & (Req.-route) Please set the quickest route to go. \\ 
        \bottomrule
        \toprule
        \multicolumn{2}{l}{\ac{dd}} \\
        \midrule
        \multirow{4}{*}{\bf Ctx.} 
        & \textit{A:} Hi. Have you got a personal computer? \\
        & \textit{B:} Certainly. What ' s the matter? \\
        & \textit{A:} I wonder if you often trade with others \\
        & on the internet. \\
        \midrule
        \multirow{2}{*}{\bf Ref.} 
        & Sure. I often buy things or do business through \\
        & it without going out to the physical stores. \\ 
        \midrule
        {\bf Pred.}
        & Yes, but I think it is a little different way. \\ 
        \bottomrule
    \end{tabular}
\end{table}

\paragraph{Sentence sentiment control}

Finally, we inspect the capability of our model for controllable generation on Yelp reviews, pre-processed by \citep{li2018delete}. The Yelp dataset is of larger scale, containing 180,000 negative reviews and 270,000 positive ones. For a controllable generation process, the symbolic vector $\y$ is provided to guide the sampling in latent space; see details in \cref{app:extra_details}. Following \citet{pang2021latent}, we train the model with sentiment supervision and use the same pre-trained classifier to determine the sentiment of the generated sentence. The pre-trained classifier has an accuracy of $98.5\%$ on the testing data and thus can accurately evaluate the sentiment of given sentences. The quantitative and qualitative results are summarized in \cref{tab:yelp_sent_control,tab:yelp_samples}, respectively. \ac{ldebm} generates positive and negative reviews with a nearly saturate accuracy, significantly outperforming all the baselines.

\begin{table}[ht!]
    \caption{\textbf{Accuracy of sentence attribute control on Yelp.}}
    \label{tab:yelp_sent_control}
    \vskip 0.1in
    \centering
    \small
    \begin{sc}
        \begin{tabular}{lccc}
            \toprule
            Model & {\bf Overall$^\uparrow$} & {\bf Positive$^\uparrow$} & {\bf Negative$^\uparrow$}  \\
            \midrule
            { DGM-VAE $+ \II$} & 64.7\% & 95.3\% &  34.0\%  \\
            { CGAN} & 76.8\% & 94.9\% &  58.6\%  \\
            { \ac{svebm}-IB} & \underline{90.1\%} & \underline{95.1\%} & \underline{85.2\%} \\
            \midrule
            \rowcolor{gray}
            { Ours} & \textbf{99.0\%} & \textbf{98.8\%} & \textbf{99.1\%}  \\
            \bottomrule
        \end{tabular}
    \end{sc}
\end{table}

\begin{table}[ht!]
    \caption{\textbf{Generated positive and negative reviews on Yelp.}}
    \label{tab:yelp_samples}
    \vskip 0.1in
    \centering
    \small
    \begin{tabular}{ll}
        \toprule
        \multirow{7}{*}{\bf Positive}
        & The food here was very tasty and \\
        & our server was very attentive. \\
        \cmidrule(l){2-2}
        & I was very satisfied for my birthday party! \\
        \cmidrule(l){2-2}
        & Definitely the best Philly Cheesesteaks \\
        & I've ever been. \\
        \cmidrule(l){2-2}
        & They are the best customer service ever! \\
        \midrule
        \multirow{6}{*}{\bf Negative}
        & Ugh the staff is so incompetent and rude. \\
        \cmidrule(l){2-2}
        & It just can't make it worse. \\
        \cmidrule(l){2-2}
        & Avoid this company at all costs. \\
        \cmidrule(l){2-2}
        & Just ruined the experience with a horrible \\ 
        & attitude on it. \\
        \bottomrule
    \end{tabular}
\end{table}

\subsection{Semi-supervised Classification}

In this experiment, we switch from neural sequence models used in previous experiments to neural document models \citep{miao2016neural,card2018neural}; we show our model can be similarly extended to semi-supervised settings as in \citet{pang2021latent} and benefit from the better learned latent space. Our model is evaluated on AGNews \citep{zhang2015character}, a popular benchmark for text classification with 127,600 documents from 4 classes. \cref{tab:ssl} shows that \ac{ldebm} performs the best when having only partial access to ground-truth data labels; it further validates the proposed formation for learning a well-structured latent space.

\begin{table}[ht]
    \caption{\textbf{Accuracy on AGNews.} We report semi-supervised classification accuracy with varied number of labeled data.}
    \label{tab:ssl}
    \vskip 0.1in
    \centering
    \small
    \begin{sc}
        \begin{tabular}{lcccc}
            \toprule
            Model  & {200} & {500} & {2500} & {10000} \\
            \midrule
            { Glove-ID} & 70.4 & 78.0 & 84.1 & 87.1 \\ 
            { Glove-OD} & 68.8 & 78.8 & 85.3 & 88.0 \\
            { VAMPIRE} & 82.9 & 84.5 & 85.8 & 87.7 \\
            { Hard EM} & 83.9 & 84.6 & 85.1 & 86.9 \\
            { CatVAE} & 84.6 & 85.7 & 86.3 & 87.5 \\
            { \ac{svebm}} & 84.5 & 84.7 & 86.0 & 88.1 \\
            { \ac{svebm}-IB} & \underline{86.4} & \underline{87.4} & \underline{87.9} & \underline{88.6} \\
            \midrule
            \rowcolor{gray}
            { Ours} & \textbf{87.4} & \bf{88.1} & \textbf{89.2} & \textbf{90.1} \\
            \bottomrule
        \end{tabular}
    \end{sc}
\end{table}

\section{Discussions and Related Work}\label{sec:related_work}

\paragraph{Text modeling}

\ac{vae} has been one of the most prominent latent variable models for generative modeling \citep{kingma2013auto,rezende2014stochastic}. It is first applied to text modeling in \citet{bowman2016generating}, followed by a wide range of work attacking challenging text generation problems using the shared framework of \ac{vae}. These include dialog generation \citep{serban2016building,serban2017hierarchical,wen2017latent,zhao2017learning,zhao2018unsupervised,fang2019implicit}, machine translation \citep{zhang2016variational}, text summarization \citep{li2017deep}, and paraphrase generation \citep{gupta2018deep}. In parallel, extensive efforts have been made to address issues like posterior collapse \citep{bowman2016generating,higgins2016beta,zhao2017learning,zhao2018adversarially,he2018lagging,li2019surprisingly,fu2019cyclical} and mode-collapse \citep{shi2020dispersed} in training \ac{vae} to further improve the language modeling performance and text generation quality. 

The interpretability of the generation process is naturally brought up as the generation quality achieves impressive progress. Recently, \citet{zhao2018unsupervised}, \citet{shi2020dispersed}, and \citet{pang2021latent} have explored interpretable text generation with deliberately designed latent spaces. \citet{zhao2018unsupervised} use a discrete latent space to capture dialog actions; \citet{shi2020dispersed} adopt a mixture of Gaussians as the \ac{vae} prior. To further improve the expressivity of latent space, \citet{pang2021latent} propose a symbol-vector coupling energy-based prior to learn a structured latent space. The coupling formulation provides a natural interface to induce the symbolic representation, which  eliminates the need of training extra auxiliary inference networks for symbol induction. Our formulation inherits the advantages from \citet{pang2021latent} by choosing an appropriate symbol-vector coupling scheme and principally incorporating the \ac{ib}. We further develop a geometric clustering-based regularization that complements the \ac{ib}; it alleviates the mode-collapse problem in variational learning of the latent space model.

\paragraph{Energy-based model}

\acp{ebm} \citep{xie2016theory,nijkamp2019learning,nijkamp2020anatomy,han2020joint} have drawn growing interest in generative modeling. As an interesting branch, \citet{pang2020learning} learn an \ac{ebm} in the latent space as a prior model for continuous latent variables; it greatly improves the expressivity over non-informative priors and demonstrates strong performance on downstream tasks, \eg, image segmentation, molecule generation, and trajectory prediction \citep{yu2021unsupervised,pang2020molecule,pang2021trajectory,jing2019task,jing2018learning}. However, both \ac{ebm} and latent space \ac{ebm} require \ac{mcmc} sampling to learn the model. The degenerate sampling quality in practice can lead to poor generation quality and instability in training \citep{grathwohl2019your,du2020improved}. We leverage diffusion models as a cure for the vanilla latent space \ac{ebm} in this work; the proposed model shows reliable sampling quality in practice.

\paragraph{Diffusion model}

Diffusion models \citep{sohl2015deep,ho2020denoising,gao2020learning}, originating from \citet{sohl2015deep}, learn from a sequence of noise-perturbed versions of the data. From such perturbed data, one can learn the conditional model to invert the diffusion process and generate high-quality samples given noisy inputs. On another front, \citet{song2019generative,song2020improved,song2020score} extend the denoising score matching method \citep{vincent2011connection}, modeling the diffusion process with continuous time step. Our formulation moves the model to the latent space in a variational framework with two benefits: (a) learning in a lower-dimensional space enables faster sampling and better convergence, and (b) learning the diffusion model in a continuous latent space avoids the discreteness of text data, which hinders the direct application of vanilla diffusion models to text modeling \citep{austin2021structured}.

Similar to our work, \citet{wehenkel2021diffusion}, \citet{sinha2021d2c}, \citet{nie2021controllable}, and \citet{vahdat2021score} have proposed to learn a diffusion model in the latent space. Specifically, \citet{wehenkel2021diffusion} empirically demonstrate that a diffusion prior can perform better than the non-informative Gaussian prior when jointly trained with a \ac{vae}. \citet{sinha2021d2c} combine contrastive learning with diffusion models in the latent space of \acp{vae} for controllable generation. \citet{nie2021controllable} and \citet{vahdat2021score} extend the idea of \citet{song2020score} in the latent space: \citet{nie2021controllable} perform controllable image generation by training a latent energy-based attribute classifier on a pre-trained generator; \citet{vahdat2021score} train score-based denoising diffusion models in the latent space of a powerful \ac{vae} \citep{vahdat2020NVAE}. Both methods have achieved very impressive image generation results. However, the listed methods are generally limited to image generation with tailored or pre-trained encoders and decoders. In contrast, our method is a general improvement for the sampling quality of latent space \ac{ebm}; it is not restricted to a certain data type. Moreover, the proposed model can be trained from scratch to form a well-structured latent space, in contrast to \citet{vahdat2021score} and \citet{nie2021controllable} which require a pre-learned latent space.

\section{Conclusion and Future Works}

We presented \ac{ldebm}, a novel symbiosis between symbol-vector coupling \ac{ebm} and diffusion model that offers the best of both worlds. The proposed model shows reliable sampling quality, learns a well-structured and meaningful latent space from scratch, and can be flexibly extended to scenarios where data labels are available. It demonstrates superior performance over strong baselines on interpretable text modeling. We hope our work inspires future research along this challenging but promising research direction. A potential follow-up research problem is to reuse powerful pre-trained language models. One could consider integrating pre-trained models with our method to realize high-quality controllable generation at low computational cost.

\textbf{Acknowledgements: }
Y. N. Wu was supported by NSF DMS-2015577. We would like to thank the anonymous reviewers for their constructive comments.

\bibliography{main}
\bibliographystyle{icml2022}

\clearpage
\renewcommand\thefigure{A\arabic{figure}}
\setcounter{figure}{0}
\renewcommand\thetable{A\arabic{table}}
\setcounter{table}{0}
\renewcommand\theequation{A\arabic{equation}}
\setcounter{equation}{0}
\pagenumbering{arabic}
\renewcommand*{\thepage}{A\arabic{page}}
\appendix

\section{Extended Derivations and Further Discussion}

\subsection{Derivation of Conditional \texorpdfstring{\acp{ebm}}{}}\label{app:cond_ebm}

We first define the marginal \acp{ebm} at each diffusion step:
\begin{equation}
    \left\{
    \begin{aligned}
        & p_\alpha(\z_t) = \frac{1}{Z_{\alpha, t}} \exp(F_\alpha(\z_t, t)) p_0(\z_t),~t=T-1 \\
        & p_\alpha(\z_t) = \frac{1}{Z_{\alpha, t}} \exp(F_\alpha(\z_t, t)),~t=0,1,...,T-2
    \end{aligned}
    \right.
    \label{equ:app_marg_ebm}
\end{equation}
where the marginal energy term is in a log-sum-exponential form 
$
    F_\alpha(\z_t, t) = \log \sum_\y \exp(\langle \y, f_\alpha(\z_t, t)\rangle)
$; it serves to aggregate the energy score from each category. Of note, the marginal \ac{ebm} corresponding with the last diffusion step has a slightly different definition. We set this term as exponential tilting of a non-informative Gaussian prior $p_0(\z_t)$ which helps to stabilize training in practice.

Recall that $\z_{t+1} = \sqrt{1 - \sigma^2_{t+1}}\z_t + \sigma_{t+1}\bm{\epsilon}_{t+1}$. Let $\tilde{\z}_t = \sqrt{1 - \sigma^2_{t+1}}\z_t$. For $t=0,1,...,T-2$, we have
\begin{equation}
    \begin{aligned}
        & p_\alpha( \tilde{\z}_t|\z_{t+1} )
                = \frac{p_\alpha(\tilde{\z}_t)p(\z_{t+1}|\tilde{\z}_t)}
                       {p_\alpha(\z_{t+1})} \\
        &= \frac{1}{\Tilde{Z}_{\alpha, t}} 
           \frac{\exp(F_\alpha(\tilde{\z}_t, t))}{p_\alpha(\z_{t+1})}
           \exp{\left(
                -\frac{1}{2\sigma^2_{t+1}}||\tilde{\z}_t - \z_{t+1}||^2
           \right)} \\
        &= \frac{1}{\Tilde{Z}_{\alpha, t}(\z_{t+1})} 
                \exp{
                \left(
                    F_\alpha(\tilde{\z}_t, t) - 
                    \frac{1}{2\sigma^2_{t+1}}||\tilde{\z}_t - \z_{t+1}||^2
                \right)},
    \end{aligned}
\end{equation}
where $\Tilde{Z}_{\alpha, t} = (2\pi\sigma^2_{t+1})^\frac{n}{2} Z_{\alpha, t}$; we slightly abuse the notation and use $p(\z_{t+1} | \tilde{\z}_t)$ to represent the forward transition $q(\z_{t+1}|\z_t)$ defined in \cref{equ:forward} for notation consistency. 

The diffused samples at time step $T$ are close to Gaussian white noise; $p_\alpha(\tilde{\z}_{T-1}|\z_T)$ therefore falls to its marginal distribution $p(\tilde{\z}_{T-1})$ defined in \cref{equ:app_marg_ebm}.

\subsection{Derivation of the ELBO}\label{app:elbo}

Recall that the ELBO in \ac{svebm} is
\begin{equation}
    \begin{aligned}
    &\ELBO_{\theta, \phi} 
          = \log p_\theta(\x) - \KL( q_\phi(\z|\x) \| p_\theta(\z|\x) ) \\
         &= \E_{q_\phi(\z|\x)} [ \log p_\beta(\x|\z) ] 
          - \KL( q_\phi(\z|\x) \| p_\alpha(\z) ) \\
         &= \E_{q_\phi (\z|\x)} \left[ 
                \log p_\beta(\x|\z) - 
                \log q_\phi (\z|\x) +
                \log p_\alpha(\z)
            \right], \\
    \end{aligned}
\end{equation}
where $\KL$ denotes the Kullback-Leibler divergence. Let us  consider the full trajectory of the perturbed samples $\z_0, \z_1, ..., \z_T$. The above equation can be written as 
\begin{equation}
    \begin{aligned}
        \ELBO_{\theta, \phi} &= \E_{q_\phi (\z_0|\x)} \left[ 
                   \log p_\beta(\x|\z_0) - 
                   \log q_\phi (\z_0|\x) 
                \right]\\
            &+ \E_{q_\phi (\z_0|\x)}
            \left[
                \log \int_{\z_{1:T}} p_\alpha(\z_{0:T}) d\z_{1:T}
            \right],
    \end{aligned}
\end{equation}
where the last term is further lower-bounded by introducing the forward trajectory distribution; the inequality holds by applying Jensen's Inequality:
\begin{equation}
    \begin{aligned}
          &\E_{q_\phi (\z_0|\x)}
            \left[
                \log \int_{\z_{1:T}} p_\alpha(\z_{0:T}) d\z_{1:T}
            \right] \\
        = \quad & \E_{q_\phi (\z_0|\x)}
            \left[
                \log \int_{\z_{1:T}} 
                    q(\z_{1:T}|\z_0) 
                    \frac{p_\alpha(\z_{0:T})}
                         {q(\z_{1:T}|\z_0)} 
                    d\z_{1:T}
            \right] \\
      \ge \quad & \E_{q_\phi (\z_0|\x)}
            \left[
                \int_{\z_{1:T}} 
                    q(\z_{1:T}|\z_0) 
                    \log 
                    \frac{p_\alpha(\z_{0:T})}
                         {q(\z_{1:T}|\z_0)} 
                    d\z_{1:T}
            \right] \\
       = \quad & \E_{q_\phi (\z_0|\x), q(\z_{1:T}|\z_0)}
            \left[
                \log 
                \frac{p_\alpha(\z_{0:T})}
                     {q(\z_{1:T}|\z_0)} 
            \right].
    \end{aligned}
\end{equation}
Further, we can decompose the joint distribution of forward and backward trajectories as
\begin{equation}
    \begin{aligned}
        & \E_{q_\phi (\z_0|\x), q(\z_{1:T}|\z_0)} 
        \left[ 
            \log \frac{ p_\alpha(\z_{0:T})}{q (\z_{1:T}|\z_0)} 
        \right] = \\ 
        & \E_{q_\phi (\z_0|x), q(\z_{1:T}|\z_0)}
          \left[ 
            \log p (\z_T) + 
            \sum_{t=0}^{T-1} \log \frac{p_\alpha (\z_{t}|\z_{t+1})}
                                      {q (\z_{t+1}|\z_{t})}
          \right] = \\
        & \E \left[ 
                \log p (\z_T) + 
                \sum_{t=0}^{T-1} \log p_\alpha (\z_{t}|\z_{t+1})
             \right] + 
          \sum_{t=0}^{T-1} \HH(\z_{t+1}|\z_{t}),
    \end{aligned}
    \label{equ:app_cross_ent}
\end{equation}
where $p(\z_T)$ is standard Gaussian distribution; $\E$ is the abbreviation of $\E_{q_\phi (\z_0|x), q(\z_{1:T}|\z_0)}$. $\HH(\z_{t+1}|\z_{t}), t=0,...,1$ is the conditional entropy under the forward trajectory distribution. We obtain $\z_t$ by sampling $\tilde{\z}_t$ from $p_\alpha(\tilde{\z}_t|\z_{t+1})$ and then applying $\z_t = \tilde{\z}_t / \sqrt{1 - \sigma^2_{t+1}}$; the reverse trajectory in our model is primarily defined by $p_\alpha( \tilde{\z}_t|\z_{t+1} )$ for $t > 0$. We use $[\z_t|\z_{t+1}]$ to represent this process in the following sections; we may interchangeably use the notation of $\tilde{\z}_t$ and $\z_t$ for simplicity.

Note that the entropies can be analytically computed and do not involve learnable parameters. The joint training of inference, prior and generation models can be largely reduced to finding the agreement of the forward and reverse Markov transitions defined by $q_\phi$ and $p_\theta$ respectively.

\subsection{Detailed Discussion of Symbol Coupling}\label{app:symb_coup}

In \cref{sec:svebm}, we briefly describe how to introduce the symbolic one-hot vector $\y$. Since only $\z_0$ is connected with $\y$, we can first define the joint prior $p_\alpha(\y, \z_0)$ as in \cref{equ:app_marg_ebm} by substituting $F_\alpha(\tilde{\z}_0, 0)$ with $\langle \y, f_\alpha(\tilde{\z}_0, 0) \rangle$. Then the conditional symbol-vector coupling joint distribution follows as
\begin{equation}
    \begin{aligned}
        p_{\alpha}(\y, \z_0 | \z_1) 
            = \frac{1}{\Tilde{Z}_{\alpha,t=0}} 
                &\exp{\left(
                    \langle \y, f_\alpha(\tilde{\z}_0, 0) \rangle
                 \right)} \\
                &\exp{\left(
                    - \frac{1}{2\sigma^2_{1}}
                    ||\tilde{\z}_0 - \z_{1}||^2
                \right)}.
    \end{aligned}
\end{equation}
Note that $p_{\alpha}(\y, \z_0|\z_1)=p_\alpha(\y|\z_0) p_\alpha(\z_0|\z_1)$, \ie, $\z_0$ is sufficient for inferring $\y$ in this formulation:
\begin{equation}
    \begin{aligned}
        p_\alpha(\y | \z_0, \z_1) &= \frac{p_{\alpha}(\y, \z_0|\z_1)}
                                         {p_\alpha(\z_0|\z_1)} \\
            &= \frac{\exp{\left(
                        \langle \y, f_\alpha(\tilde{\z}_0, 0) \rangle
                     \right)}}
                    {\exp{\left(
                        F_\alpha(\tilde{\z}_0, 0)
                     \right)}},
    \end{aligned}
\end{equation}
so that given $\z_0$,
\begin{equation}
    p_\alpha(\y|\z_0) \propto 
        \exp(\langle \y, f_\alpha(\tilde{\z}_0, 0) \rangle).
    \label{equ:app_cls}
\end{equation}
It similarly becomes a softmax classifier where $f_\alpha(\tilde{\z}_0, 0)$ provides the logit scores for the $K$ categories.

\subsection{Derivation of the Information Bottleneck}\label{app:ib}

We first define the mutual information term between $\z_0$ and $\y$. Consider the joint distribution of $\x, \z_0$ and $\y$, $\pi(\y, \z_0, \x) = p_\alpha(\y|\z_0) q_\phi(\z_0|\x) q_{\rm data}(\x)$; the mutual information $\II(\z_0, \y)$ defined under $\pi$ then follows as:
\begin{equation}
\label{eq:app_mi}
    \begin{aligned}
        \II(\z_0, \y) &= \HH(\y) - \HH(\y|\z_0) \\
                      &= -\sum_\y q(\y) \log q(\y) \\
                      &\quad + \E_{q_\phi(\z_0)} \sum_\y p_\alpha(\y|\z_0) 
                               \log p_\alpha(\y|\z_0),  
    \end{aligned}
\end{equation}
where $q(\y) = \E_{q_\phi(\z_0)} [p_\alpha(\y|\z_0)]$; $p_\alpha(\y|\z_0)$ is the softmax probability over $K$ categories in \cref{equ:app_cls}. 

We then show how to obtain the quantities defined in \cref{sec:ib}. For the marginal distribution of $\z_0$: 
\begin{equation}
    \begin{aligned}
        q_\phi(\z_0) &= \int_{\x, \z_{1:T}} Q_\phi(\x, \z_{0:T}) d\x d\z_{1:T} \\
                     &= \E_{q_{\rm data}(\x)} [q_\phi(\z_0 | \x)].
    \end{aligned}
\end{equation}

The entropy and conditional entropy of $\z_0$ are thus
\begin{equation}
    \begin{aligned}
    & \HH(\z_0) = -\E_{q_\phi(\z_0)}[\log q_\phi(\z_0)]; \\
    & \HH(\z_0|\x) = - \E_{Q_\phi(\x, \z_0)}[\log q_\phi(\z_0|\x)].
    \end{aligned}    
\end{equation}

Taking together, we can then decompose the KL-Divergence, $\KL( Q_\phi \| P_\theta )$, in \cref{equ:kld_joint} as:
\begin{equation}
    \begin{aligned}
         \KL( Q_\phi \| P_\theta )
          &= \E_{Q_\phi} \left[ q_{\rm data}(\x) \right] 
           + \E_{Q_\phi} \left[ q_\phi(\z_{0:T} | x) \right] \\
          &- \E_{Q_\phi} \left[ p_\alpha(\z_{0:T}) \right]
           - \E_{Q_\phi} \left[ p_\beta(\x | \z_{0}) \right],
    \end{aligned}
\end{equation}
and further as:
\begin{equation}
    \label{equ:app_kld_dec}
    \begin{aligned}
         - \HH(\x) &+ \sum_{t=0}^{T-1} \HH(\z_{t+1}|\z_{t}) - \HH(\z_0|\x) 
                    + {\color{blue} \HH(\z_0) - \HH(\z_0)} \\
                   &- \E_{Q_\phi} \left[ p_\alpha(\z_{0:T}) \right]
                    - \E_{Q_\phi} \left[ p_\beta(\x | \z_{0}) \right], 
    \end{aligned}
\end{equation}
by plugging in $\HH(\z_0) - \HH(\z_0) = 0$. Rearranging \cref{equ:app_kld_dec}, we can obtain
\begin{equation}
    \begin{aligned}
         \KL( Q_\phi \| P_\theta ) &= \mathcal{C} 
                 - \E_{Q_\phi} \left[ p_\beta(\x | \z_{0}) \right]  \\
                &+ \KL(q_\phi(\z_{0}) \| p_\alpha(\z_{0:T})) 
                 + \II(\x, \z_0),
    \end{aligned}
\end{equation}
which leads to our result in \cref{equ:kld_mi}.

\subsection{Derivation of the Learning Gradient}\label{app:grad}

Recall that we derive the extended version of \cref{equ:elbo_svebm} in \cref{app:elbo}. To calculate the gradient of $\alpha$, we have
\begin{equation}
    \begin{aligned}
        \nabla_\alpha \ELBO_{{\rm Diff}, \theta, \phi} &=
            \nabla_\alpha \E 
             \left[
                \sum_{t=0}^{T-1} \log p_\alpha (\z_{t}|\z_{t+1})
             \right] \\
        &= \E \left[
                \sum_{t=0}^{T-1} \nabla_\alpha 
                    \log p_\alpha (\z_{t}|\z_{t+1})
             \right],
    \end{aligned}
\end{equation}
where $\E$ is the abbreviation of $\E_{q_\phi (\z_0|x), q(\z_{1:T}|\z_0)}$; in practice, we use Monte-Carlo average to approximate the expectation. We next examine the learning gradient for each diffusion step $t$.
\begin{equation}
    \begin{aligned}
        \nabla_\alpha \log p_\alpha (\z_{t}|\z_{t+1}) = 
            \nabla_\alpha F_\alpha(\tilde{\z}_t, t) - 
            \nabla_\alpha \Tilde{Z}_{\alpha, t}(\z_{t+1}),
    \end{aligned}
\end{equation}
where the quadratic term $\frac{1}{2\sigma^2_{t+1}}||\tilde{\z}_t - \z_{t+1}||^2$ is not related to $\alpha$ and gets cancelled. According to the definition of the partition function in \cref{sec:svebm}, we can similarly derive
\begin{equation}
    \begin{aligned}
        \nabla_\alpha \Tilde{Z}_{\alpha, t}(\z_{t+1}) = 
            \E_{p_\alpha(\tilde{\z}_t|\z_{t+1})} \left[
                \nabla_\alpha F_\alpha(\tilde{\z}_t, t)
            \right],
    \end{aligned}
\end{equation}
as in \citet{pang2020learning}. For the prior model, we thus have
\begin{equation}
    \begin{aligned}
        \nabla_\alpha \ELBO_t &= 
             \E_{q_\phi(\tilde{\z}_t, \z_0|\x)} 
                [\nabla_\alpha F_\alpha(\tilde{\z}_t, t)] \\
          &- \E_{q_\phi(\z_{t+1}, \z_0|\x), p_\alpha(\tilde{\z}_t|\z_{t+1})}
                [\nabla_\alpha F_\alpha(\tilde{\z}_t, t)],
    \end{aligned}
\end{equation}
where $q_\phi(\tilde{\z}_t, \z_0|x) = q(\tilde{\z}_t|\z_0) q_\phi(\z_0|\x)$. Note that we can sample $\z_t,~t>0$ directly from 
\begin{equation}
    q(\z_t|\z_0) = \N(\z_{t}; 
                \sqrt{\Bar{\gamma_t}}\z_{t-1}, 
                (1 - \Bar{\gamma_t}) \I),
\end{equation}
by merging the Gaussian noises during forward diffusion process; we denote $\gamma_t = 1-\sigma^2_{t}$ and $\Bar{\gamma_t} = \prod_{i=1}^t \gamma_t$.

For the encoder and decoder, based on \cref{equ:elbo_svebm} and \cref{equ:app_cross_ent}, we have 
\begin{equation}
    \begin{aligned}
    &\nabla_\psi \ELBO 
        = \nabla_\psi \E_{q_\phi(\z_0|\x)} [
            \log p_\beta(\x|\z_0) - \log q_\phi (\z_0|\x)
        ] \\
    &   -\nabla_\phi \E_{q_\phi (\z_{0:T}|\x)} 
             \left[ 
                \log p (\z_T) + 
                \sum_{t=0}^{T-1} \log p_\alpha (\z_{t}|\z_{t+1})
             \right],
    \end{aligned}
\end{equation}
where the summation of energy terms provides extra guidance for the optimization of encoder.

\section{Extra Experiment Details and Discussion}

\subsection{Network Architecture and Hyperparameters}\label{app:impl}

\begin{table}[ht!]
    \caption{\textbf{Network architecture for the \ac{ldebm} prior.} $N$ is set to $12$ for all the experiments.}
    \label{tab:app_arch} 
    \vskip 0.1in
    \centering
    \small
    \begin{tabular}{ccc}
        \toprule
        {\bf Layers} & {\bf Output size} & {\bf Note} \\
        \cmidrule(r){1-1} \cmidrule(r){2-2} \cmidrule(r){3-3}
        
        \multicolumn{3}{c}{\bf Time Embedding} \\
        \hline
        Input: $t$ & $1$ & \tabincell{c}{Index of \\ diffusion step} \\
        Sin. embedding & $200$ & {} \\
        Linear, LReLU & $200$ & \tabincell{c}{negative\_slope \\ $0.2$} \\
        Linear & $200$ & {} \\
        
        \hline
        \multicolumn{3}{c}{\bf Input Embedding} \\
        \hline
        Input: $\z$ & $d_{\text{lat}}$ & {} \\
        Linear, LReLU & $200$ & \tabincell{c}{negative\_slope \\ $0.2$} \\
        Linear & $200$ & {} \\
        
        \hline
        \multicolumn{3}{c}{\tabincell{c}{{\bf Context Embedding} \\ (for response generation only)}} \\
        \hline
        Input: $\z_{\text{ctx}}$ & $512$ & \tabincell{c}{ctx. embedding} \\
        Linear, LReLU & $200$ & \tabincell{c}{negative\_slope \\ $0.2$} \\
        Linear & $200$ & {} \\
        
        \hline
        \multicolumn{3}{c}{\bf \ac{ldebm} Prior} \\
        \hline
        Input: \tabincell{c}{$\z, t$ \\ $^*\z_{\text{ctx}}$} & \tabincell{c}{$1$, $d_{\text{lat}}$ \\ $512$} & optional $\z_{\text{ctx}}$ \\
        Embedding & $200$ & \tabincell{c}{Embedding of \\ each input} \\
        Concatenate & \tabincell{c}{$400$ \\ $600$} & \tabincell{c}{w/o ctx. \\ w/ ctx.} \\
        LReLU, Linear & $200$ & \tabincell{c}{negative\_slope \\ $0.2$} \\
        N ResBlocks & $200$ & \tabincell{c}{LReLU, Linear \\ $+$ Input} \\
        LReLU, Linear & $K$ & $K$ class logits \\
        Log-Sum-Exp & $1$ & energy score \\
        \bottomrule
    \end{tabular}
\end{table}

We provide detailed network architecture for the latent space model of this work in \cref{tab:app_arch}; we adopt the same architecture throughout the experiments. Spectral normalization \citep{miyato2018spectral} is used to regularize parameters in linear layers. The encoder and decoder in all models are the same as in \citet{pang2021latent}, implemented with a single-layer GRU with a hidden size of $512$. The key hyperparameters of \ac{ldebm} for each dataset are listed in \cref{tab:app_hypp}. Of note, we use the same dimension of the latent space as in \citep{pang2021latent} for a fair comparison.

\begin{table}[ht!]
    \caption{\textbf{Hyperparameters of \ac{ldebm}.} \textsc{\ac{dd}-CLS} presents the set of hyperparameters used in unsupervised clustering on \ac{dd} dataset. \textsc{\ac{dd}-GEN} presents the set of hyperparameters used in conditional response generation on \ac{dd} dataset. }
    \label{tab:app_hypp} 
    \vskip 0.1in
    \centering
    \small
    \begin{tabular}{l|ccccc}
        \toprule
        \textsc{Dataset} & $d_{\text{lat}}$ & $K$ & $\lambda_1$ & $\lambda_2$ & $\lambda_3$ \\
        \hline
        \textsc{2D Gaussian} & 2 & 16 & 1 & 0.05 & 0.05 \\
        \textsc{2D Pinwheel} & 2 & 10 & 1 & 0.05 & 0.05 \\
        \hline
        \textsc{\ac{ptb}} & 40 & 20 & 0.1 & 0.05 & 0.05 \\
        \textsc{Jericho} & 40 & 20 & 0.1 & 0.05 & 0.05 \\
        \textsc{\ac{dd}-CLS} & 32 & 125 & 0.01 & 0.05 & 0.5 \\
        \textsc{\ac{dd}-GEN} & 32 & 125 & 1 & 0.05 & 0.05 \\
        \textsc{\ac{smd}} & 32 & 125 & 10 & 10 & 5 \\
        \textsc{Yelp} & 40 & 2 & 50 & 50 & 200 \\
        \hline
        \textsc{AGNews} & 20 & 4 & 1e-3 & 5 & 200 \\
        \bottomrule
    \end{tabular}
\end{table}
$\lambda_1$ is the hyperparameter that reweights the term in \cref{equ:app_cross_ent}; it generally controls how fast $q_\phi$ and $p_\theta$ run towards each other. $\lambda_2$ refers to the hyperparameter in \cref{equ:kld_mi}; it controls the trade-off between the compressivity of $\z_0$ about $\x$ and its expressivity to $\y$. $\lambda_3$ controls the weight of classification loss mentioned in \cref{sec:gc}; recall that we use pseudo-label $\hat{\y}$ inferred by the geometric clustering algorithm or the ground-truth label $\y$ to supervise $p_\alpha(\y|\z_0)$ in our modeling. For controllable generation and semi-supervised classification, we find it important to have a larger weight on the classification loss so that the model is forced to capture the major modes of the data.

For optimization, we use Adam optimizer \citep{kingma2014adam} with $\beta_1=0.9$ and $\beta_2=0.999$ for all the experiments. On all the datasets but 2D synthetic datasets and AGNews dataset, we use a batch size of $128$ and a constant learning rate of $1e-3$ for encoder and decoder without weight decay. For \ac{ldebm}, we use a constant learning rate of $1e-4$. We use a larger batch size of $1000$ on 2D synthetic datasets. On the AGNews dataset, we use the same set of hyperparameters as in \citet{pang2021latent} for optimization. The batch size is set to $200$; the initial learning rate is $1e-4$ for encoder and decoder, and $1e-5$ for \ac{ldebm}. Learning rates are exponentially decayed with a decay rate of $0.998$ for each model. Encoder and \ac{ldebm} have a weight decay rate of $2e-3$ and $1e-3$, respectively.

\subsection{Experiment Settings and Baselines}\label{app:db}

\paragraph{Experiment settings} 

For generative modeling, following previous methods \citep{shi2020dispersed,pang2021latent}, the \ac{nll} term is computed with importance sampling \citep{burda2016importance} using $500$ importance samples. To compute \ac{rppl}, we set the generated sample size as $40,000$, which is the same size as \ac{ptb} training set. We recruit ASGD Weight-Dropped LSTM \citep{merity2018regularizing} to compute \ac{rppl} as in previous works. 

In terms of conditional response generation, for word-embedding-based evaluation on \ac{smd} and \ac{dd}, we use the publicly available GloVe \citep{pennington2014glove} word embeddings of $300$ dimension trained on 840B tokens, and report the score from 1 response per context. We use a context window size of 5 during training and evaluation.

The maximum length of each sentence is set to $40$ words for most datasets and $70$ words for the JerichoWorld dataset. On JerichoWorld dataset, we extract the description of each state as the text data.

\paragraph{Baselines}

On {\bf \ac{ptb}, \ac{dd}} and {\bf \ac{smd}}, our model is compared with the following baselines: (1) RNNLM \citep{mikolov2010recurrent}, the language model implemented with GRU \citep{cho2014learning}; (2) AE \citep{vincent2010stacked}, the deterministic auto-encoder which has no regularization to the latent space; (3) DAE, the AE with a discrete latent space; (4) \ac{vae} \citep{kingma2013auto}, the vanilla \ac{vae} with a continuous latent space and a non-informative Gaussian prior; (5) DVAE, the \ac{vae} with a discrete latent space; (6) DI-VAE \citep{zhao2018unsupervised}, a DVAE variant with a mutual information term between the observed piece of text $\x$ and its inferred latent variable $\z$; (7) semi-VAE \citep{kingma2014semi}, the semi-supervised \ac{vae} model with independent discrete and continuous latent variables; (8) GM-VAE, the \ac{vae} with a Gaussian mixture prior; (9) DGM-VAE \citep{shi2020dispersed}, the GM-VAE with a dispersion term that avoids the mode-collapse of Gaussian mixture prior; (10) semi-VAE $+ \II(\x,\y)$, GM-VAE $+ \II(\x,\y)$, DGM-VAE $+ \II(\x,\y)$, are the same models as (7), (8), and (9) respectively, but with a mutual information term between $\x$ and $\y$ computed using separate inference networks for $\y$ and $\z$. We compare with the close competitors (11) \ac{svebm}, the symbol-vector coupling prior model and (12) \ac{svebm}-IB, \ac{svebm} with regularization based on information-bottleneck. 

On {\bf{Yelp}} dataset, we additionally include text conditional GAN \citep{subramanian2018towards} as a baseline for controllable generation. On {\bf AGNews} dataset, we further compare our model to VAMPIRE \citep{gururangan2019variational}, a VAE-based semi-supervised text learning model. Other baselines include its supervised learning variants: (1) the model trained with Glove embedding pre-trained on $840$ billion words (Glove-OD); (2) the model trained with Glove embedding on in-domain unlabeled data (Glove-ID). We also include more recent baselines such as Hard EM and CatVAE \citep{jin2020discrete} that improve over VAMPIRE. 

\subsection{Extra Details for Experiments}\label{app:extra_details}

\paragraph{More ablation study}

We conduct additional experiments on both \acs{ptb} and \acs{dd} datasets to inspect the contribution of the proposed techniques. In \cref{sec:gen_model}, we have reported results on \acs{ptb} and  datasets of \textsc{Ours w/o GC} which represents the model with \acl{ib} but without \ac{gc}; \textsc{Ours} denotes the full model. 

We further conduct experiments on the proposed model without using \ac{ib} or \ac{gc}. We observe that the proposed model using only diffusion-based sampling scheme has a \ac{rppl} of 166.26, BLEU of 11.30, \ac{wkl} of 0.07 and \ac{nll} of 80.76 on PTB; it has a MI of 0.01, BLEU of 19.28, Act. of 0.12 and Emo. of 0.06 on \ac{dd}, which is better than \acsp{svebm} (please see \cref{tab:ptb_gen,tab:dd_cluster} in \cref{sec:gen_model}). 

We also add \ac{gc} to \acs{svebm} (denoted as \textsc{SVE-IB w/ GC}). We find that \textsc{SVE-IB w/ GC} does perform better compared with \textsc{SVE-IB}, showing a \ac{rppl} of 179.95, BLEU of 10.08, \ac{wkl} of 0.15 and \ac{nll} of 93.28 on PTB; it has a MI of 2.88, BLEU of 11.75, Act. of 0.61 and Emo. of 0.60 on \ac{dd}. Notably, \textsc{SVE-IB w/ GC} is still inferior to \acsp{ldebm}. 

In summary, we think these additional experiments (1) emphasize the importance of our diffusion-based modeling approach, and (2) demonstrate the effectiveness of \ac{gc} as additional regularization. 

\paragraph{2D synthetic data}

\begin{figure*}[!htbp]
    \centering
    \subfigure[\ac{svebm}-IB Gaussian]{
        \includegraphics[width=0.75\columnwidth]{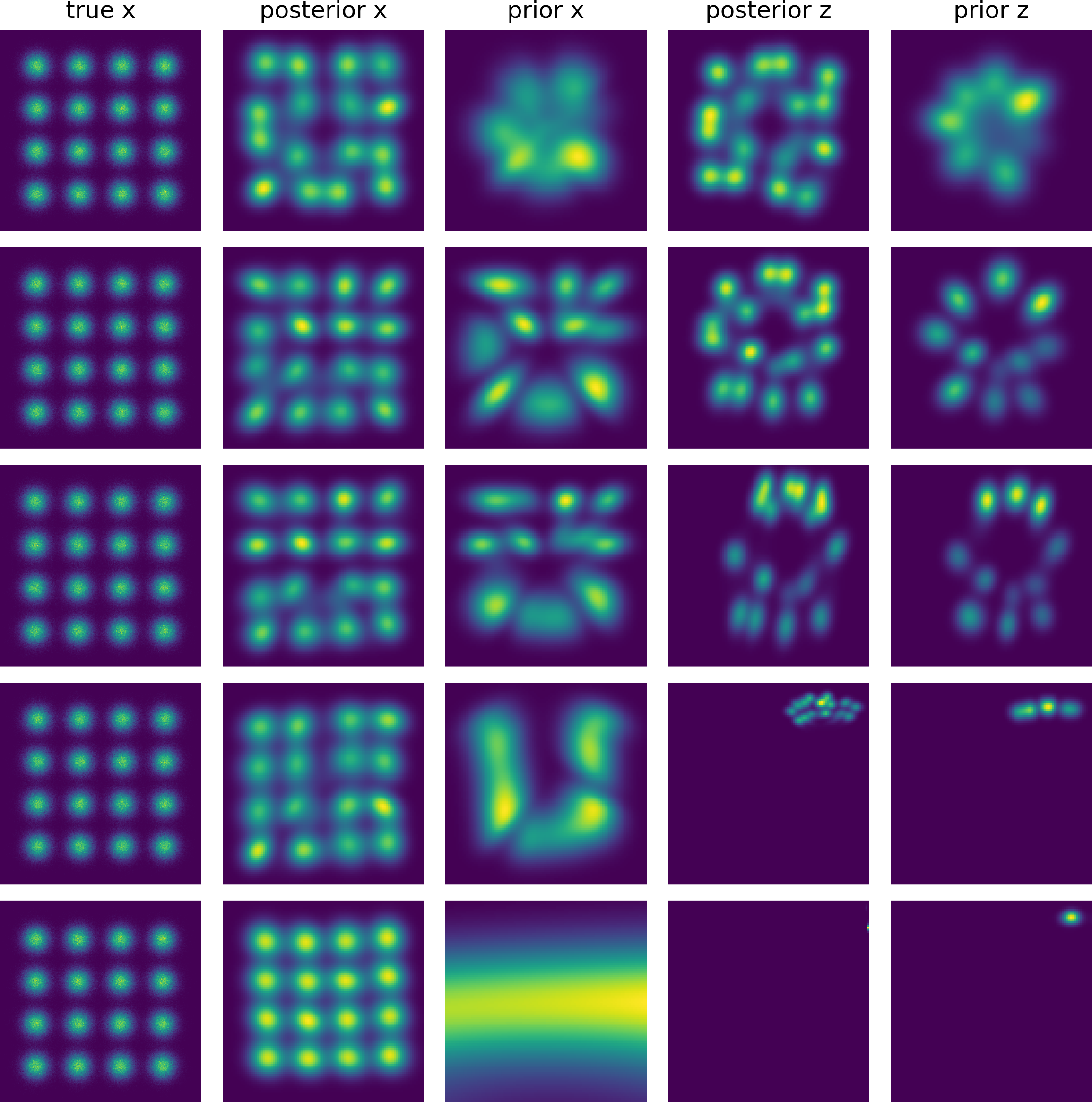}    
        \label{fig:app_baseline_gaussian}
    }
    \subfigure[\ac{svebm}-IB Pinwheel]{
        \includegraphics[width=0.75\columnwidth]{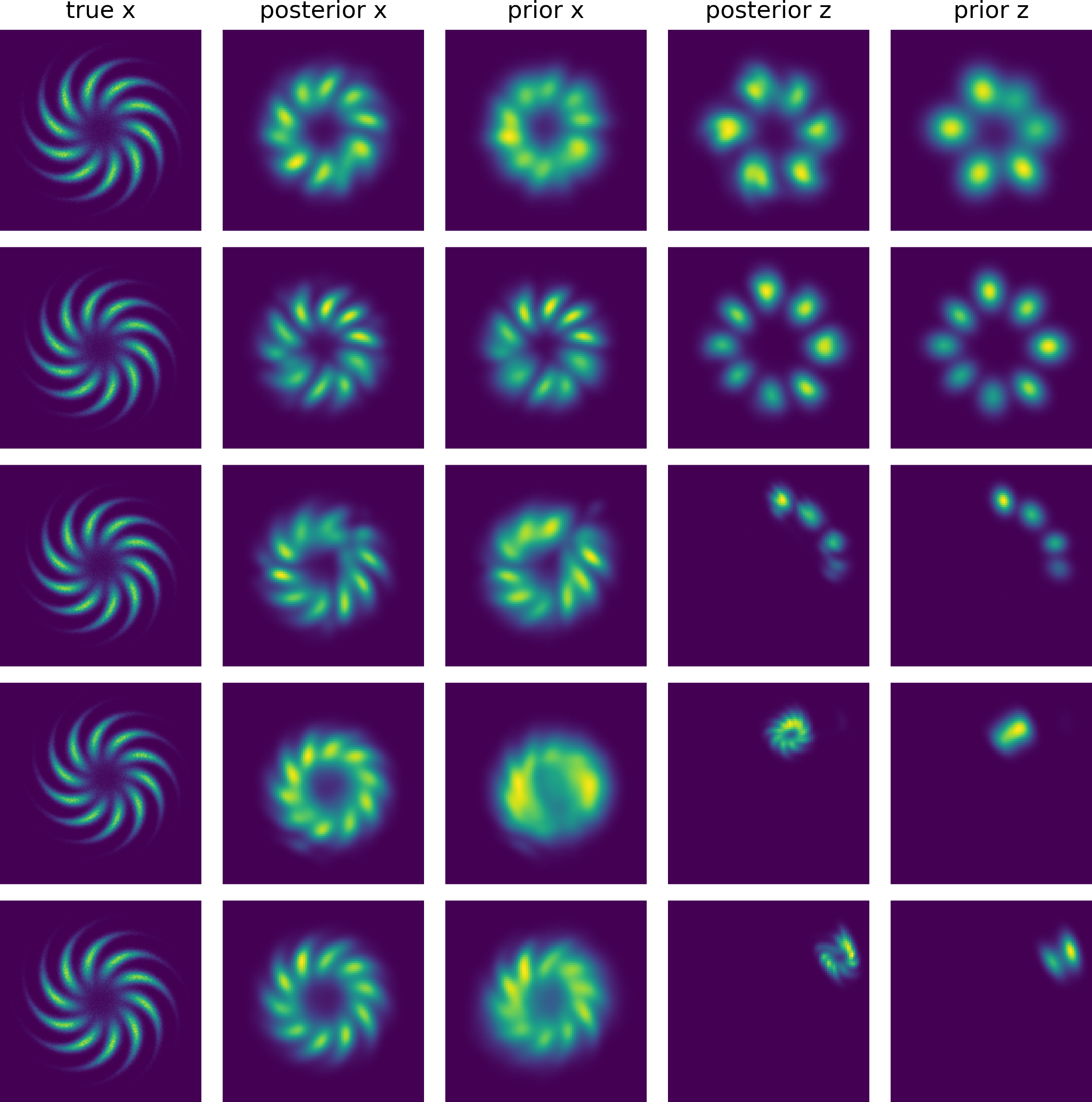}    
        \label{fig:app_baseline_pinwheel}
    }
    \subfigure[\ac{ldebm} w/o PL Gaussian]{
        \includegraphics[width=0.75\columnwidth]{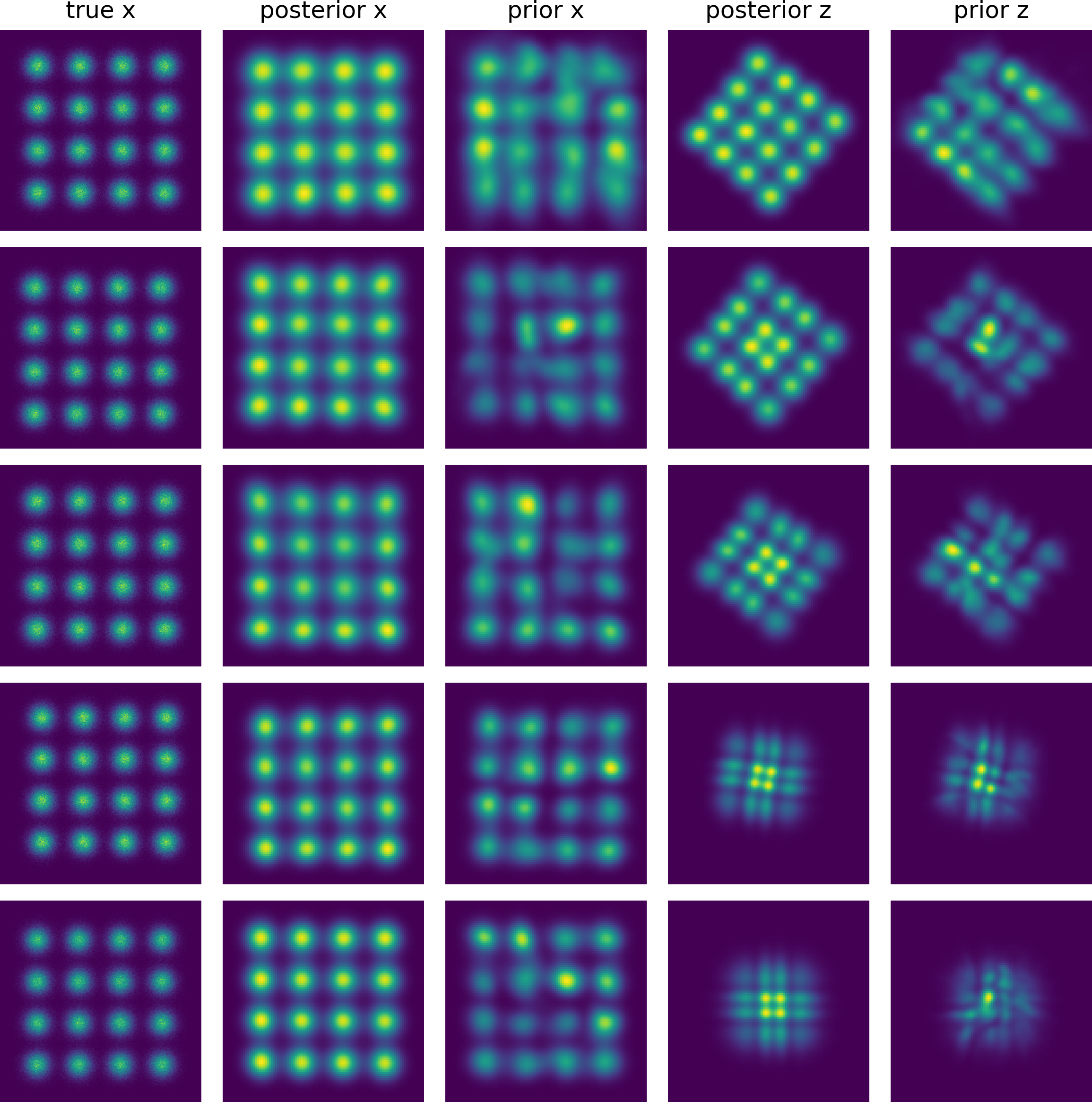}    
        \label{fig:app_baseline_nPL_gaussian}
    }
    \subfigure[\ac{ldebm} w/o PL Pinwheel]{
        \includegraphics[width=0.75\columnwidth]{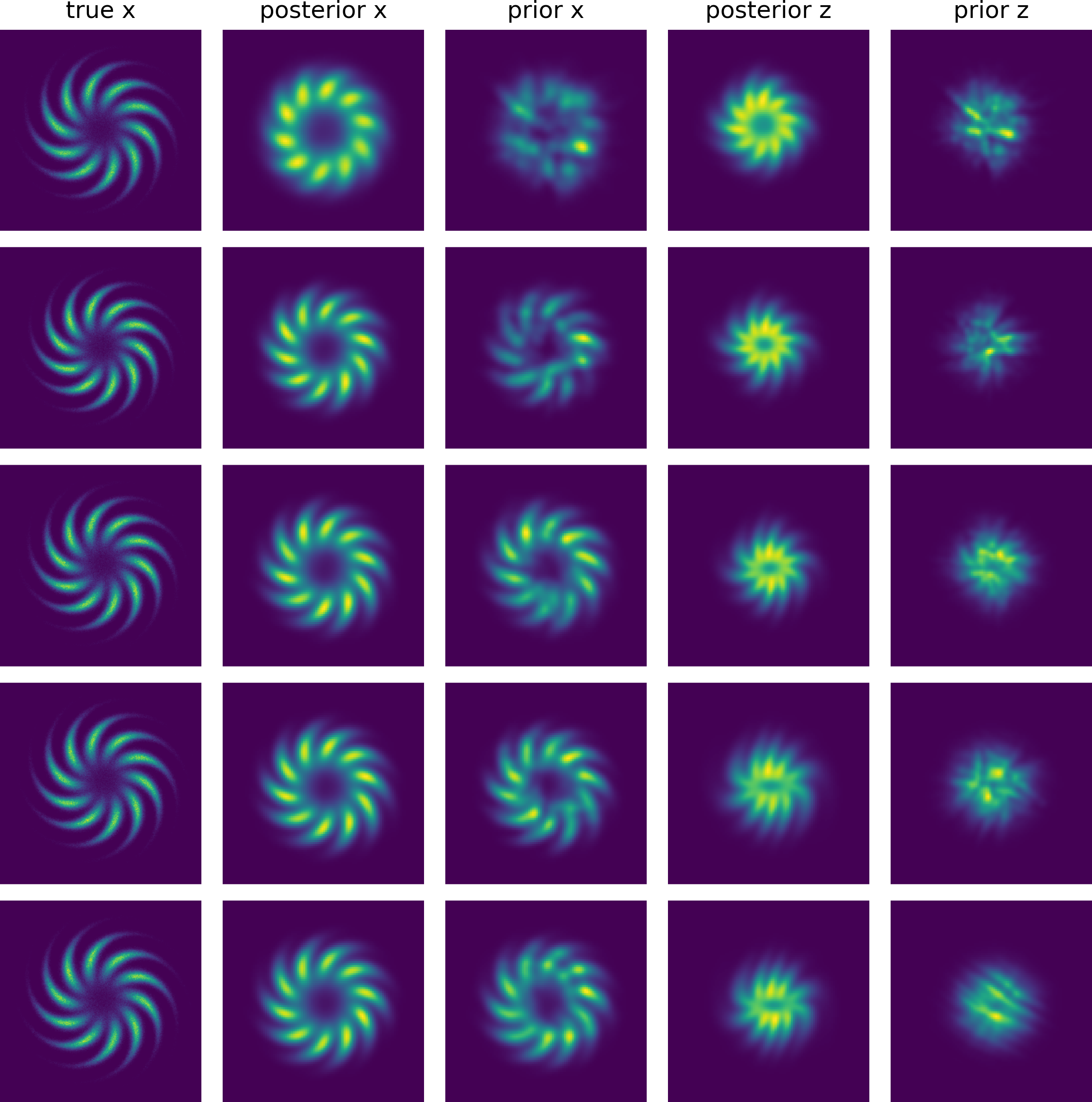}    
        \label{fig:app_baseline_nPL_pinwheel}
    }
    \subfigure[\ac{ldebm} Gaussian]{
        \includegraphics[width=0.75\columnwidth]{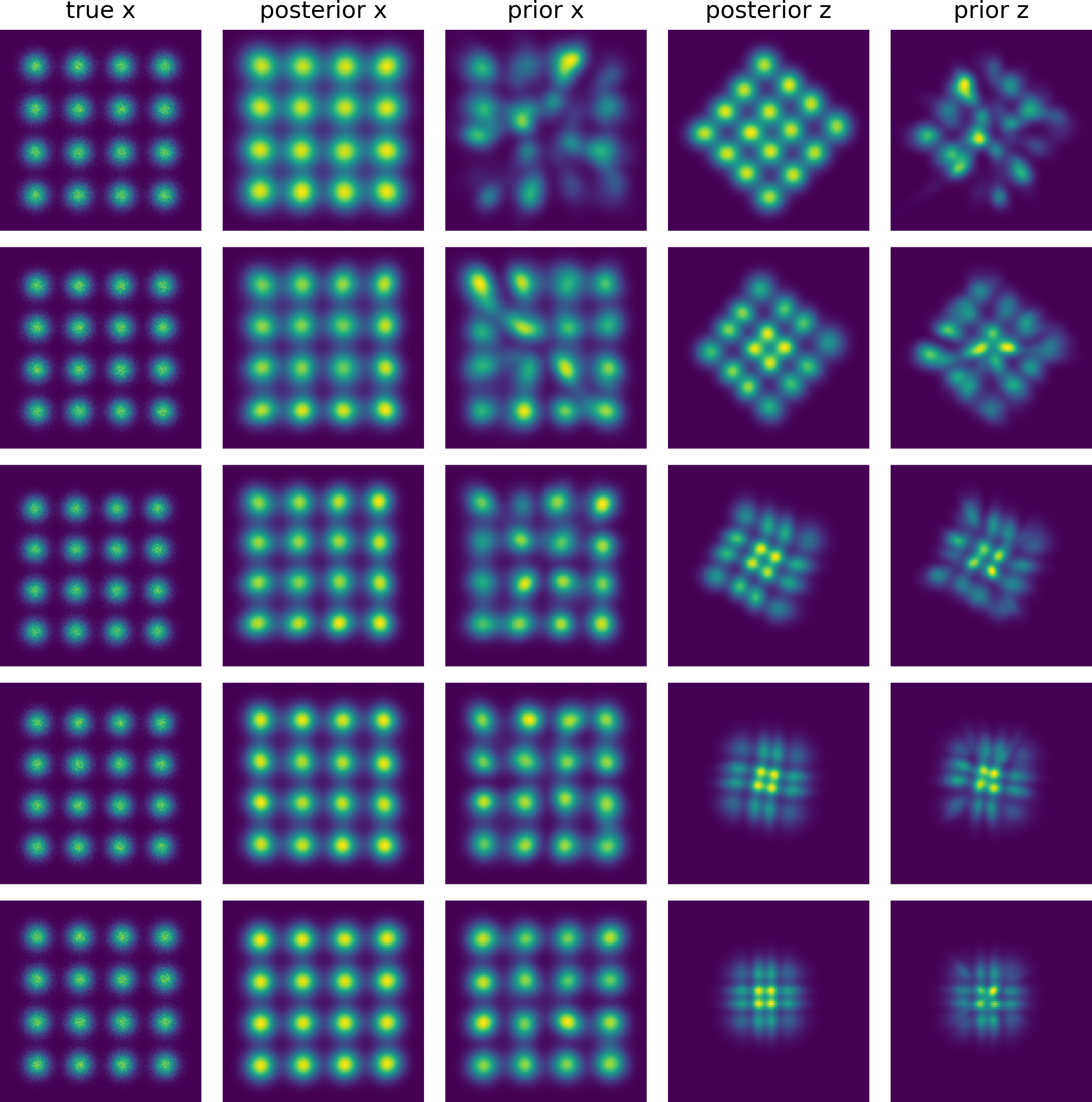}    
        \label{fig:app_gaussian}
    }
    \subfigure[\ac{ldebm} Pinwheel]{
        \includegraphics[width=0.75\columnwidth]{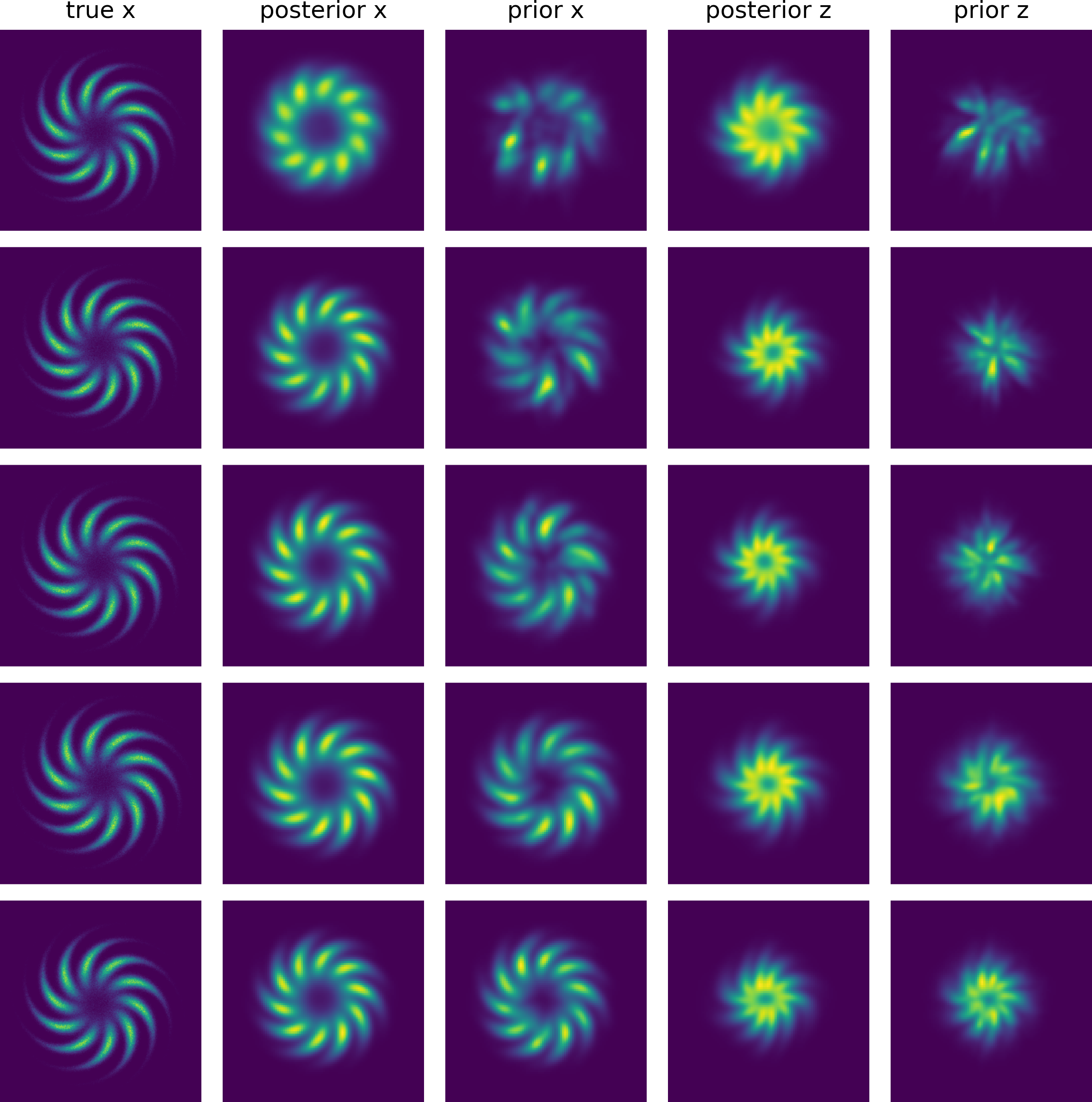}    
        \label{fig:app_pinwheel}
    }
    \caption{\textbf{Full evolution of \ac{svebm}-IB and our models.} In each sub-figure, we provide the typical states of the model trained on the corresponding dataset, sequentially from the top row to the bottom row.}
    \label{fig:app_evolve}
\end{figure*}

We provide the full evolution of \ac{svebm}-IB and our models as visualized in \cref{fig:app_evolve}. Though \ac{svebm}-IB can capture some regularities of the data in the early stages of training, the model is prone to collapse due to the degenerated sampling quality. This features an exploding KL-term and leads to poor performance on generation. Our preliminary experiments indicate that common deep learning heuristics for improving the model capacity barely help. These include but are not limited to increasing the number of parameters in \ac{svebm}, \ie, using larger models, and adopting deliberately designed activation functions or normalization modules. \ac{ldebm} w/o geometric clustering has a better sampling quality and effectively mitigates the instability in training. However, the mode coverage is not satisfying in data space; the structure is unclear in latent space. In contrast, \ac{ldebm} w/ geometric clustering shows superior generation quality with better mode coverage. It demonstrates a better-structured latent space.

\paragraph{Sentence completion}

To perform sentence completion, we adopt a two-stage training scheme. We first train the \ac{ldebm} with inference, prior and generation models on the JerichoWorld dataset. After the first-stage training, the parameters of prior, inference and generation models are fixed. We then train a shallow MLP in the latent space to project the inferred posterior $\z_0$ to a disentangled space; the variables in the projected $\z_0$ can be grouped as: (a) the representation of observable words $\hat{\z}_{\text{obs}}$ in the input sentence and (b) the representation of unknown words $\hat{\z}_{\text{unk}}$. Conditional sampling in the latent space then refers to updating $\hat{\z}_{\text{unk}}$ based on the fixed $\hat{\z}_{\text{obs}}$ by running Langevin dynamics guided by the latent space model.

We mask half of the words in the sentences with \texttt{<unk>} token to prepare the inputs. In the second stage of training, we supervise the MLP by minimizing the reconstruction error between only the observable words of the input the sentence and the corresponding outputs of the model.

\paragraph{Sentence sentiment control}

Recall that in our formulation only $\z_0$ is connected to $\y$. We therefore condition only the final reverse diffusion step $[\z_0|\z_1]$ on $\y$ when performing controllable generation, \ie, using $\y$ to guide the generation only when $t=0$ in \cref{alg:synth}. This can be a bit counter-intuitive since no label information is injected in previous reverse steps. Theoretically, $\y$ and $\z_{1:T}$ are independent given $\z_0$ in our formulation; however, we empirically observe that $\y$ and $\z_t$ for $t > 0$ are nearly independent even marginally after we integrating out $\z_{0:t-1}$ in our model.
In other words, $p_\alpha(\y|\z_t),~t>0$ are in general non-informative since adding noise in the latent space could be much more corrupting than adding noise in the data space. The model learns to enjoy the less multi-modal energy landscape in previous reverse steps; it then seeks the given mode only in the most informative final reverse step. We examine $p_\alpha(\y|\z_t),~t>0$ for the model trained on Yelp dataset by marginalizing out $\z_{t-1}$ of $p_\alpha(\y, \z_{t-1} | \z_t),~t>0$. For example, for $t=1$, we may calculate 
\begin{equation}
    \begin{aligned}
    p_\alpha(\y|\z_1) 
        &= \int_{\z_0} p_\alpha(\y|\z_0) p_\alpha(\z_0|\z_1) d \z_0 \\
        &= \E_{p_\alpha(\z_0|\z_1)} \left[ p_\alpha(\y|\z_0) \right] \\
        &\approx \frac{1}{M} \sum_{i=1}^M p_\alpha(\y|\z_0^{(i)}).
    \end{aligned}
\end{equation}

\begin{figure}[t!]
    \centering
    \includegraphics[width=0.9\linewidth]{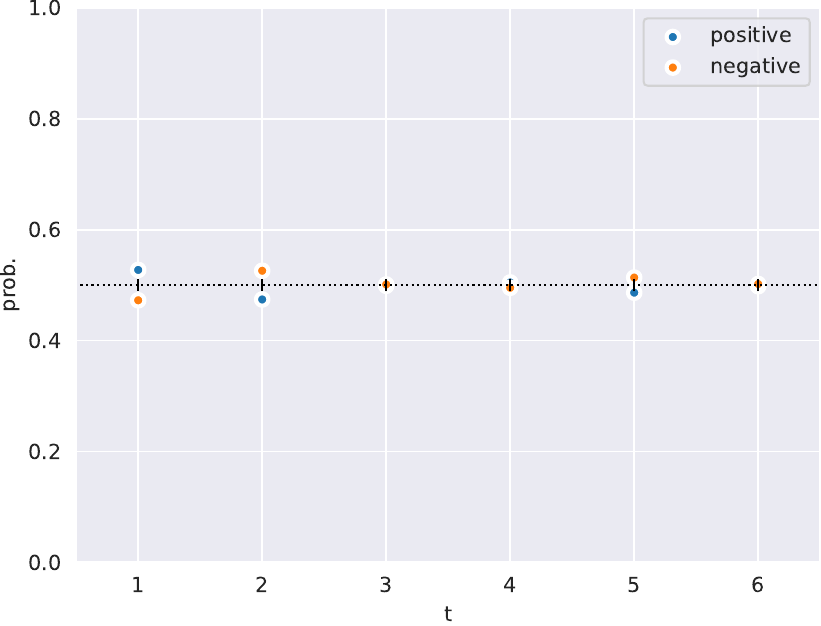}
    \caption{\textbf{Visualization of $p_\alpha(\y|\z_t)$ over $t$}. $p_\alpha(\y|\z_t)$ is constantly around the probability of $0.5$ over $t$.}
    \label{fig:app_non_info}
\end{figure}

See \cref{fig:app_non_info} for the visualization of $p_\alpha(\y|\z_t)$ over $t$. 

A more intuitive method is to use the data label $\y$ to supervise each $[\y, \z_t|\z_{t+1}]$, so that we can propagate the label information through the whole trajectory. Given $\z_0$, $\y$ and $\z_{1:T}$ are independent. But if we marginalize out $\z_0$, $\y$ will depend on $\z_1$. Similarly, if we continue to marginalize out $\z_1$, $\y$ will depend on $\z_2$. Repeating this process results in $p_\alpha(\y|\z_t)$ for each $t$ after integrating out $\z_{0:t-1}$. Supervising $p_\alpha(\y|\z_t),~t>0$ using $\y$ therefore effectively encodes the label information into the whole trajectory.

While the marginalization can be difficult, we may approximate it by learning the amortized version of $p_\alpha(\y|\z_t),~t>0$ as $p_\alpha(\y, \z_{t-1}=\mu_{\phi, t-1}|\z_t),~t>0$, where $\mu_{\phi, t}$ is the posterior mean of $\z_t$. We may therefore circumvent the intractable integration in practice and guide the whole trajectory for controllable generation.

\end{document}